\documentclass[runningheads]{llncs}

\usepackage{eccv}

\usepackage{eccvabbrv}

\usepackage{graphicx}
\usepackage{booktabs}
\usepackage{amsmath}
\usepackage{amssymb}
\usepackage{mathtools}
\usepackage{lipsum}
\usepackage{bm}
\usepackage{tabularx}
\usepackage{algorithm}
\usepackage{algpseudocode}
\usepackage{tikz}
\usetikzlibrary{calc,positioning,arrows.meta,decorations.pathreplacing}
\usepackage{xcolor}
\definecolor{pelvis}{HTML}{fd0100}
\definecolor{head}{HTML}{018000}
\definecolor{lhand}{HTML}{ffff09}
\definecolor{rhand}{HTML}{0a01fc}

\usepackage[accsupp]{axessibility}  

\usepackage{hyperref}

\usepackage{orcidlink}

\begin{document}

\title{Lifting Embodied World Models for Planning and Control} 

\titlerunning{Lifting Embodied World Models for Planning and Control}

\author{Alex N. Wang\inst{1}\and
Trevor Darrell\inst{2}\and
Pavel Izmailov\inst{1}\and 
Yutong Bai\inst{2\ddagger} \and 
Amir Bar\inst{3\ddagger}}

\authorrunning{A.~N.~Wang et al.}

\institute{
Computer Science, New York University \and
BAIR, UC Berkeley \and
Imperial College London
\footnotetext{
$^\ddagger$ Equally advised.
}
}

\maketitle
\renewcommand{\thefootnote}{}
\footnotetext{Project page: \url{https://alexn.wang/lwm}}
\renewcommand{\thefootnote}{\arabic{footnote}}

\begin{figure*}[t]
    \centering
    \includegraphics[width=\textwidth]{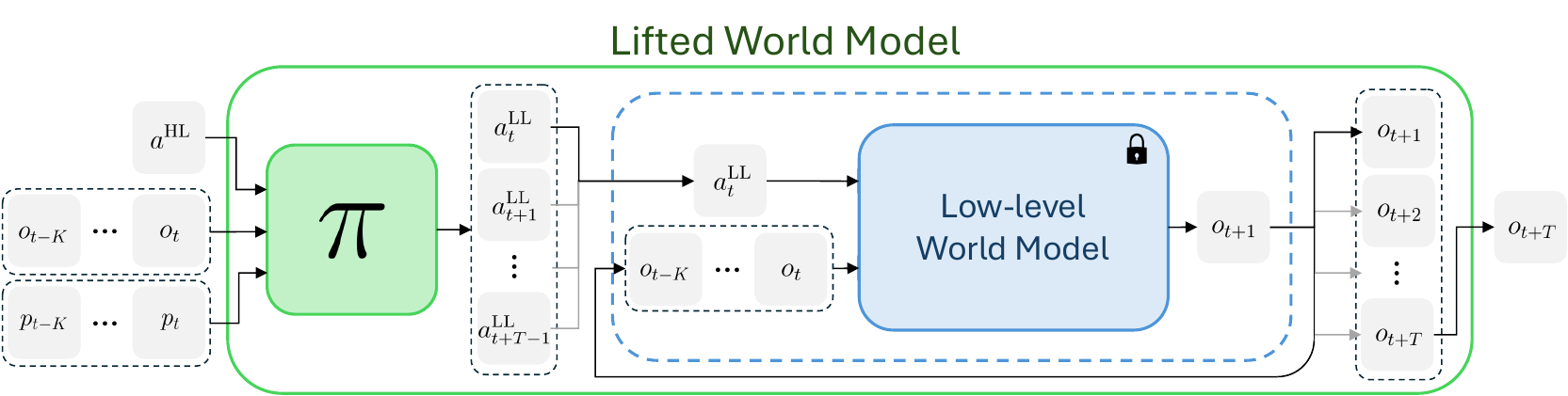}
    \caption{
    \textbf{Lifted World Model.}
    The lifted world model outputs a new observation $o_{t+T}$ given a high-level action $a^\text{HL}$. 
    First, the \textcolor[HTML]{47D45A}{policy $\pi$} predicts a sequence of $T$ low-level actions $a^\text{LL}_{t:t+T-1}$. 
    Then the \textcolor[HTML]{4E95D9}{low-level world model} autoregressively samples a sequence of new observations, one for each low-level action, appending each to the observation context for the next step.
    }
    \label{fig:teaser}
\end{figure*}

\begin{abstract}
World models of embodied agents predict future observations conditioned on an action taken by the agent. 
For complex embodiments, action spaces are high-dimensional and difficult to specify: for example, precisely controlling a human agent requires specifying the motion of each joint. 
This makes the world model hard to control and expensive to plan with as search-based methods like CEM scale poorly with action dimensionality.
To address this issue, we train a lightweight policy that maps high-level actions to sequences of low-level joint actions. Composing this policy with the frozen world model produces a \textit{lifted world model} that predicts a sequence of future observations from a single high-level action.
We instantiate this framework for a human-like embodiment, defining the high-level action space as a small set of 2D waypoints annotated on the current observation frame, each specifying a near-term goal position for a leaf joint (pelvis, head, hands).
Waypoints are low-dimensional, visually interpretable, and easy to specify manually or to search over. 
We show that the lifted world model substantially outperforms searching directly in low-level joint space ($3.8\times$ lower mean joint error to the goal pose), while remaining more compute-efficient and generalizing to environments unseen by the policy.
\end{abstract}

\section{Introduction}
World models are a promising approach to understanding the physical world from data~\cite{lecun2022path, ha2018worldmodels, mur2026v, yang2024unisim}.
They learn how actions affect the environment by predicting a future observation conditioned on a current observation and an action.
World models have progressed from narrow settings like simple games~\cite{robine2023transformer, ha2018worldmodels, hafner2025dreamerv3} to diverse environments, with actions that span text~\cite{agarwal2025cosmos}, navigation~\cite{bar2025navigationnwm, hu2023gaia}, and whole-body joint control~\cite{bai2025wholebodypeva}.

As world models target richer embodiments~\cite{hansen2024hierarchical, bai2025wholebodypeva, gao2026dreamdojo, mereu2025generative}, their action spaces have grown high-dimensional, which presents a challenge in planning and control.
For a human-like embodiment~\cite{roetenberg2009xsens}, actions are specified by high-dimensional, per-joint parameters~\cite{bjorck2025gr00t, hansen2024hierarchical, tevethumanmdm, tevet2025closd, fu2025humanplus}.
For example, reaching out with one hand requires 
coordinated extension of the shoulder, elbow, and wrist.
This high dimensionality makes world models difficult to control and especially expensive to use for planning: search-based methods like the Cross-Entropy Method (CEM)~\cite{rubinstein199789crossentropymethodcem} scale poorly in action dimensionality~\cite{bharadhwaj2020modelmpccemgradient,psenka2026parallelgrasp}.

To address this, we train a lightweight policy that maps high-level actions to sequences of low-level actions (see Figure~\ref{fig:teaser}). Composing this policy with the frozen world model yields a \textit{Lifted World Model} (LWM) that predicts a sequence of observations from a single high-level action. 
Lifting reduces the effective dimensionality and length of the action sequence the planner must search over, without requiring any changes to the world model itself.

\begin{figure}[t]
\centering
\begin{minipage}[t]{0.48\linewidth}
\centering
\includegraphics[width=0.98\linewidth,trim={.5cm 0 .48cm .58cm},clip]{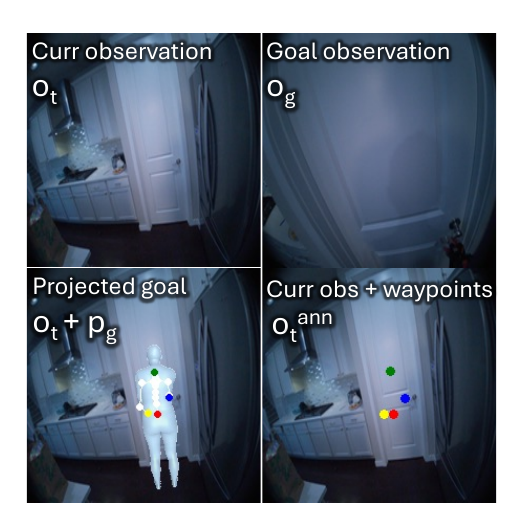}
\caption{
Specifying goals using waypoints. A subset of joints from goal pose $p_g$ is drawn as waypoints on $o_t$  to create $o_t^\text{ann}.$
}
\label{fig:waypoint_and_proprioception}
\end{minipage}
\hfill
\begin{minipage}[t]{0.48\linewidth}
\centering
\includegraphics[width=0.98\linewidth,trim={0cm 0cm 0cm 0cm},clip]{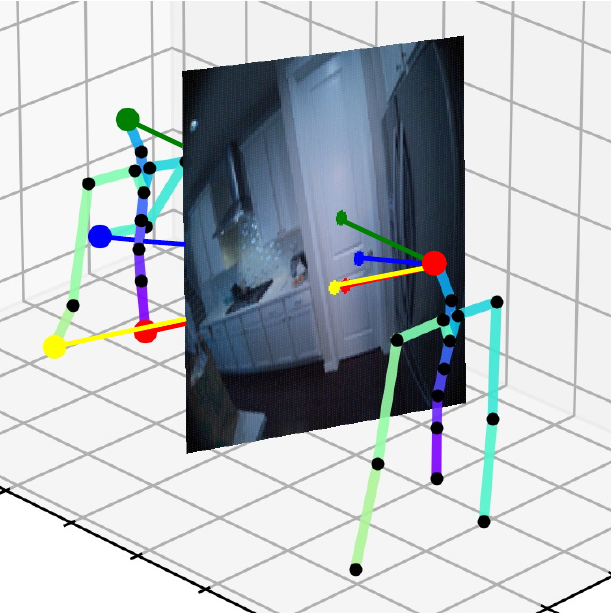}
\caption{To obtain waypoints, the goal pose $p_g$ is projected onto current observation $o_t$ seen by pose $p_t$.}
\label{fig:waypoint_generation}
\end{minipage}
\end{figure}

\begin{figure}[t]
    \centering
    \includegraphics[width=\linewidth]{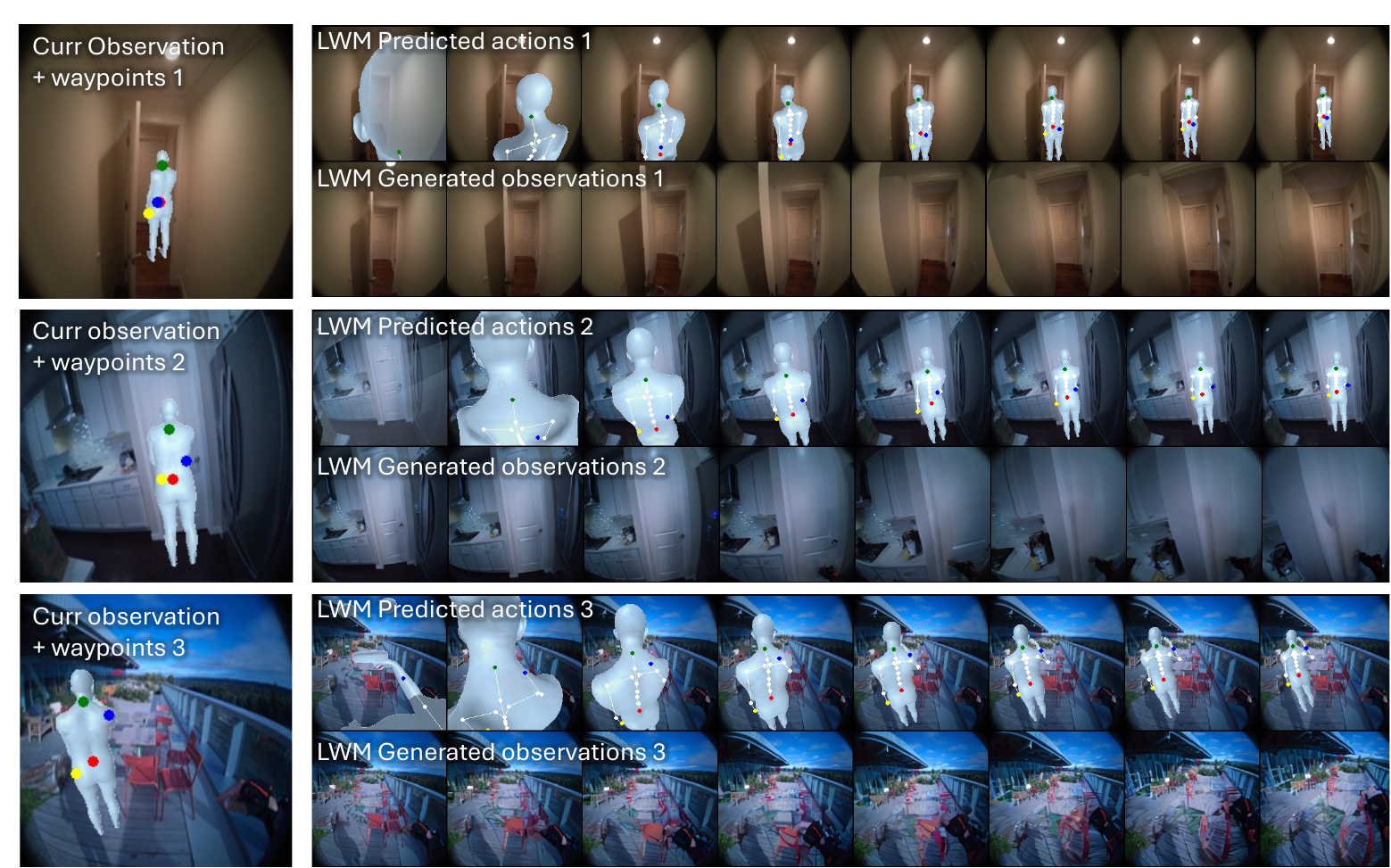}
    \caption{
    Visualization of LWM rollouts and paired actions\protect\footnotemark. 
    Left$\rightarrow$right on each row: the input goal, the sequence of low-level actions predicted by the policy (upper row), and the observations generated by the world model (lower row).
    The goal is shown by waypoints (pelvis \protect\tikz\fill[pelvis] (0,0) circle (0.6ex);, head \protect\tikz\fill[head] (0,0) circle (0.6ex);, left-hand \protect\tikz\fill[lhand] (0,0) circle (0.6ex);, right-hand \protect\tikz\fill[rhand] (0,0) circle (0.6ex);) plotted on current observation $o_t$, where each dot represents the goal-position for that joint; the mesh is shown for visualization only.
    \textbf{Ex 1:} advancing through the doorway. 
    \textbf{Ex 2:} agent walking up and grasping the doorknob. 
    \textbf{Ex 3:} avoiding obstacles and moving forward holding a camera.
    }
    \label{fig:lwm_visualizations}
\end{figure}

In our experiments, we lift PEVA\cite{bai2025wholebodypeva}, a world model that predicts egocentric observations conditioned on human joint-angle displacements. Inspired by human visuomotor control, where visuospatial goals in the posterior parietal cortex are translated into joint-level commands by the motor cortex~\cite{merel2019motorneurohierarchy}, we design our high-level action space around visual goals expressed in the agent's current view. A waypoint is the 2D projection of a 3D goal position for a leaf joint of the embodiment (pelvis, head, or hand) that can be reached in a short sequence of joint-space actions (Figure~\ref{fig:waypoint_and_proprioception}). At training time, waypoints are obtained from ground-truth future poses via 3D$\rightarrow$2D camera projection (Figure~\ref{fig:waypoint_generation}); at inference time, they can be specified manually or searched over with CEM.
Figure~\ref{fig:lwm_visualizations} shows example LWM rollouts: given waypoints overlaid on the current frame, the policy generates a sequence of joint actions and the world model rolls them forward into future observations.

In Section~\ref{sec:goal_conditioned_policy} we show that waypoints are an effective goal-conditioning signal for generating joint actions, while goal observations are not: in egocentric video the agent's own body is mostly out of frame, so a goal image carries little information about the target pose, and conditioning on it barely improves over an unconditioned policy. Waypoints, by contrast, express goals directly in the input frame and meaningfully steer the generated actions toward the target. 
We further find that the policy responds sensibly to sparse and manually specified waypoints (Figure~\ref{fig:counterfactual}) and uses image context to interpret the same waypoints differently depending on the scene (Figure~\ref{fig:goal_reuse}).

In Section \ref{sec:search-based_lwm_planning} we use the LWM for search-based planning with CEM on hybrid navigation and interaction tasks from the Nymeria dataset~\cite{ma2024nymeria}.
We find that searching in waypoint space yields solutions $3.8\times$ closer to the goal in mean joint error than searching directly in low-level joint space. Planning with high-level actions is more compute-efficient, performs better on long horizon tasks, and generalizes to environments not seen by the policy during training.

In summary, our contributions are:
\begin{enumerate}
    \item We propose a method for lifting a low-level world model to a higher abstraction level by training a lightweight policy that maps high-level actions to sequences of low-level actions, without modifying the base world model.
    \item We instantiate this framework for a human-like embodiment with a new high-level action space of waypoints: 2D image-space goal positions for the agent's leaf joints, which are low-dimensional, visually interpretable, and well-suited to search.
    \item We show that waypoints are a more effective goal-conditioning signal for egocentric policies on a human-like embodiment than goal observations.
    \item Using the resulting Lifted World Model, we improve CEM planning by 3.8× in mean joint error to the goal compared to CEM in joint space, at lower compute cost, and with generalization to environments unseen by the policy.
\end{enumerate}

\section{Preliminaries}
Here we define world models and the representations used for poses and actions.
\footnotetext{We render the mesh for illustration purposes only. The overlaid skeleton shows the exact pose of the embodied agent.}

\vspace{-0.3cm}
\subsubsection{World Models.} 
A world model predicts the future observation $o_{t+1}$ given an action $a_t$ and a current observation $o_t$. A model may condition on $K$ previous observations, $\bm{o}_t = \{o_{t-K},\dots, o_t\}$; we refer to $K$ as the \textit{world model context length}. 
The world model prediction is given by:
\begin{align}
    o_{t+1} = f_\phi(\bm{o}_t, a_t).
\end{align}
\paragraph{Notation.} We use bold symbols (e.g., $\bm{o}_t, \bm{p}_t$) to denote sequences ending at time $t$, and non-bold symbols (e.g., $o_t, p_t$) to denote individual elements at time $t$.
We also use non-bold symbols for vector-valued quantities at a single timestep (e.g., a 3D position $x$ or a vector of Euler angles $\phi$).

\vspace{-0.3cm}
\subsubsection{Pose representation.}
The human-like embodiment is represented using the XSens~\cite{roetenberg2009xsens} model following PEVA~\cite{bai2025wholebodypeva}. 
A pose $p$ consists of the 3D position $x_\text{pelvis}$ of the pelvis and Euler angles $\phi_{(\cdot)}$ for each joint.
In our experiments we use the upper-body PEVA checkpoint with 15 joints and the pose representation:
\begin{equation}
\begin{aligned}
    p = [x_\text{pelvis}^\top | \phi^\top_\text{pelvis} | \phi^\top_\text{lumbar5}|\dots|\phi^\top_\text{left\_hand}] \in \mathbb{R}^{48}.
\end{aligned}
\end{equation}
The pelvis position $x_\text{pelvis}$ localizes the pose in a reference frame while the Euler angles $\phi_{(\cdot)}$ 
describe the orientation of each joint. The model assumes rigid bones of constant length.

A single pose $p_t$ is defined with respect to its own pelvis frame; therefore at start time $t$ the position and Euler angles of the pelvis are both zero: $
x_\text{pelvis} , \phi_\text{pelvis} = \bm 0$. 
In contrast, a sequence of poses $p_t, p_{t+1}\dots$ are defined with respect to the pelvis frame of the first pose of the sequence.

\vspace{-0.3cm}
\subsubsection{Actions.}
An action $a_t$ is the change between poses $p_t, p_{t+1}$:
\begin{equation}
\begin{aligned}
    a_t &= [\delta x_\text{pelvis}^\top | \delta\phi^\top_\text{pelvis} | \delta \phi^\top_\text{lumbar5}|\dots|\delta\phi^\top_\text{left\_hand}] \in \mathbb{R}^{48}.
\end{aligned}
\end{equation}
Non-pelvis joints are defined in the frame of their parent joint, so only angular displacements are needed; their 3D positions are recovered via forward kinematics
(Appendix \ref{appendix:forward_kinematics}).
\begin{align}
    \delta x_\text{pelvis} &= x_{\text{pelvis},t+1} - x_{\text{pelvis},t}, \\
    \delta \phi_{(\cdot)} &=\mathrm{R}^{-1}(\mathrm{R}(\phi_{t,(\cdot)})^\top\mathrm{R}(\phi_{t+1,(\cdot)}) ).
\end{align}
$\mathrm{R}$ and $\mathrm{R}^{-1}$ transform Euler angles to and from rotation matrices in SO(3).

\section{Lifting an Ego World Model}
Here we introduce the three components of our method.
In Section~\ref{sec:high_level_action_space}, we describe the high-level waypoint action space for controlling egocentric, embodied agents.
In Section~\ref{sec:high_level_action_policy}, we present the waypoint-conditioned policy that predicts sequences of low-level joint actions.
Finally, in Section~\ref{sec:lifted_world_model_method}, we combine the policy with a low-level world model to form a\textit{ Lifted World Model} (LWM) and describe its use in planning.

\subsection{High-level Action Space}
\label{sec:high_level_action_space}
The high-level action space should be effective, low-dimensional and intuitive.
We express high-level actions $a^\text{HL}$ in the current observation $o_t$ as a set of \textit{waypoints}.
A waypoint is the 2D projection into the current observation $o_t$ of a near-term 3D goal position for one leaf joint of the embodied agent.
We incorporate them by annotating $o_t$ with each waypoint to obtain $o_t^\text{ann}$ (see Figure~\ref{fig:waypoint_and_proprioception}).

Defining high-level actions as 2D points in $o_t$ makes them visually interpretable and easy to specify manually at test time.
Even more importantly, because waypoints are finite and low dimensional, they avoid the exponential scaling with action dimensionality in search-based planning such as CEM.
Our method is in contrast to specifying goals using a goal image $o_g$ as in NoMaD~\cite{sridhar2024nomad}.
Their approach requires the availability of goal images while also being high dimensional and 
ill-suited to search.

\vspace{-0.3cm}
\subsubsection{Waypoints for a Human Embodiment.}
For the XSens embodiment used by PEVA, we define $a^\text{HL}$ using four waypoints --- for the pelvis, head, left and right hands:
\begin{align}
    a^\text{HL} = \{w_\text{pelvis}, w_\text{head}, w_\text{left\_hand}, w_\text{right\_hand}\}.
\end{align} 
These are the leaf joints in the embodiment, affording navigation, interaction and camera control (see Figure~\ref{fig:waypoint_and_proprioception}).
When plotting the waypoints on image $o_t$, we differentiate them by assigning each a unique color: pelvis \tikz\fill[pelvis] (0,0) circle (0.6ex);, head \tikz\fill[head] (0,0) circle (0.6ex);, left hand\tikz\fill[lhand] (0,0) circle (0.6ex);, right hand \tikz\fill[rhand] (0,0) circle (0.6ex);.
While 2D image points do not fully specify a 3D goal, the image $o_t$ and the spacing between waypoints provide context for inferring the goal pose.

\vspace{-0.3cm}
\subsubsection{Computing Ground-truth Waypoints.}
During training, high-level action labels are computed by projecting the ground-truth goal pose $p_g$ onto the current image $o_t$.
3D positions are computed from the pose representation with forward kinematics (see Appendix~\ref{appendix:forward_kinematics}), which are then projected into the image plane with the camera matrix $\mathbf{P}$ (Figure~\ref{fig:waypoint_generation})
\begin{align}
    \{x_\text{pelvis}, \dots, x_\text{left\_hand}\} &= \text{forward\_kinematics}(p_g), \\
    w_{(\cdot)} &= \mathbf{P}x_{(\cdot)}.\footnotemark
\end{align} \footnotetext{Homogeneous coordinates and the camera frame transform are omitted for clarity.}

\vspace{-1em}
\subsection{High-level Action Conditioned Policy}
\label{sec:high_level_action_policy}
We train a lightweight policy to map high-level actions into the low-level action space. 
At time $t$ the policy predicts a sequence of low-level actions $a^\text{LL}_{t:t+T-1}\in\mathbb{R}^{T\times 48}$ conditioned on context observations $\bm{o}_t = \{o_{t-K_\pi}, \dots, o_t\}\in[0,1]^{(K_\pi+1)\times CHW}$, poses $\bm{p}_t = \{p_{t-K_\pi}, \dots, p_t\}\in \mathbb{R}^{(K_\pi+1)\times48}$, and high-level action $a^\text{HL}_t\in[0,1]^8$:
\vspace{-0.5em}
\begin{align}
    a^\text{LL}_{t:t+T-1} \sim \pi_\theta(\bm{o}_t, \bm{p}_t, a^\text{HL}_t).
\end{align}
We allow for sparse actions where waypoints are out-of-frame, requiring the policy to infer the final joint positions. 
Goal masking is used to hide goal conditioning to generate unconditioned actions $p(a^\text{LL}_{t:t+T-1}|\bm{o}_t, \bm{p}_t)$.
The policy context size $K_\pi$ can be different from the world model context $K$.
By training a policy, we capture short-term motion patterns like locomotion, reaching and grasping. 

\vspace{-0.8em}
\subsubsection{Goal masking.}
We explore removing goal-masking when training our final lifting policy to prioritize goal-conditioned generation.
When trained without masking, our policy can generate unconditioned actions by using a goal image $o_g$ without any waypoints.
We explore waypoint masking in Appendix~\ref{appendix:waypoint_masking}.

\vspace{-0.3cm}
\subsubsection{Architecture.}
The policy $\pi(\cdot)$ is a diffusion policy that generates actions by denoising in joint action space. 
First, a goal image $o_t^\text{ann}$ is created by annotating each waypoint onto observation $o_t$. 
Then, images $\bm o_t$ and $o_t^\text{ann}$ are encoded using a DINOv3-S~\cite{simeoni2025dinov3} encoder. 
The tokens are not pooled to preserve spatial information, producing embeddings $[z_{t-K_\pi:t}| z_t^\text{ann}] \in \mathbb{R}^{(K_\pi+2)\times L \times D}$.
Context poses $\bm p_t$ are linearly projected into $D$ dimensions and added to the image embeddings of the corresponding timestep: $z_{(\cdot)}' = z_{(\cdot)} + \text{proj}(p_{(\cdot)})$. 
Additional 3D positional embeddings are added to tokens to preserve spatial as well as temporal ordering before processing with a vision transformer.
The processed tokens are pooled to form context vector $c_t\in \mathbb{R}^D$, which is used as input to condition the denoising UNet during generation.

Our method adds pose context, a DINOv3 encoder, later spatial pooling, 3D positional embeddings, and a larger denoising network to a base diffusion policy~\cite{sridhar2024nomad, chi2025diffusionpolicy}.
See Appendix~\ref{appendix:policy_architecture} for an architecture figure.

\subsection{Lifted World Model}
\label{sec:lifted_world_model_method}
Using the policy, we can lift the low-level world model to use inputs from the high-level action space (Figure~\ref{fig:teaser}).
The \textit{Lifted World Model} (LWM) predicts a future observation $o_{t+T}$ from image, pose context, and a single high-level action:
\begin{align}
    o_{t+T} = f^\text{HL}(\bm{o}_t, \bm{p_t}, a^\text{HL}_t).
\end{align}
Internally, the policy predicts a sequence of low-level actions that are used to autoregressively sample a sequence of internal observations $o_{t+1:t+T-1}$ from the low-level world model.
\begin{align}
    a^\text{LL}_{t:t+T-1} &= \pi_\theta(\bm o_t, \bm p_t, a^\text{HL}_t) \\
    o_{\tau+1} &= f_\phi(\bm o_\tau, a^\text{LL}_\tau)
\end{align}
for $\tau \in \{t, \dots, t+T-1\}$. 
Generated observations are appended to the observation context $\bm o_\tau$ for subsequent steps $\tau > t$.

\vspace{-0.3cm}
\subsubsection{Search-based Planning.}
\begin{figure*}[t]
    \centering
    \includegraphics[width=0.75\linewidth]{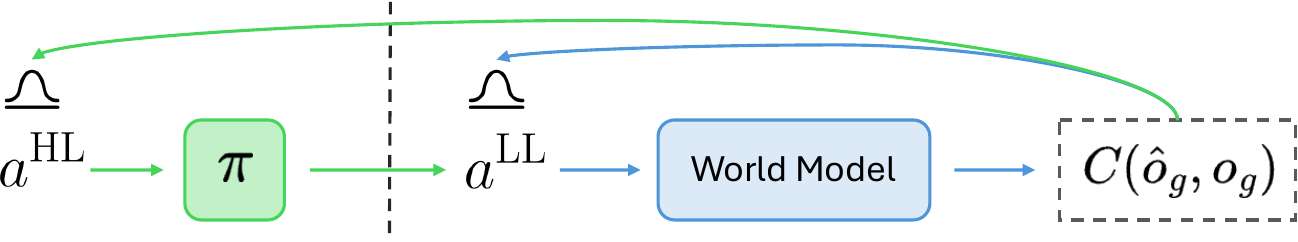}
    \caption{
    Planning using the \textcolor[HTML]{47D45A}{Lifted World Model}.
    The action prior, sampling, and updates to the distribution occur in high-level action space.
    In contrast, planning with the \textcolor[HTML]{4E95D9}{low-level world model} operates in the low-level action space.
    }
    \label{fig:planning_pipeline}
\end{figure*}
Now with the LWM, we can run cross-entropy method (CEM) in the high-level action space.
Given a starting state and a goal, we sample a sequence of actions, simulate future observations using the world model and select the trajectory that arrives closest to the goal.
Using a low-level world model, actions are sampled in the low-level space directly.
The number of samples required for effective search grows
exponentially with action dimensionality times sequence length.
With a lifted world model, actions can be sampled in the smaller (8 vs. 48 dimensions per step), high-level action space $a^\text{HL}$, greatly reducing both dimensionality and sequence length.
The selected high-level action is then mapped to low-level joint actions via the policy for execution.
See Figure~\ref{fig:planning_pipeline} for a visualization.

Formally, with a goal observation $o_g$, perceptual similarity cost function $C(\cdot)$, and a distribution over low-level actions $p_\text{LL}(\cdot)$, the CEM planning objective can be formulated as: 
\begin{align}
    \bm a^{\text{LL}*} = \underset{\bm a^\text{LL} = a_t,\dots,a_{t+T-1}}{\arg\min} \, \mathbb{E}_{\hat{o}_g\sim f_\phi(\bm a^\text{LL})}
    [C(\hat{o}_g, o_g)],
\end{align}
while planning with a lifted world model uses the high-level prior $p_\text{HL}(\cdot)$ and is written as
\begin{align}
    \bm a^{\text{HL}*} = \underset{\bm a^\text{HL}=a^\text{HL}_t,\dots,a_{t+(L-1)T}^\text{HL}}{\arg\min} \, \mathbb{E}_{\hat{o}_g\sim f_\phi(\bm{a}^\text{LL}), \bm{a}^\text{LL}\sim\pi_\theta(\bm a^\text{HL})
    }[C(\hat{o}_g, o_g)].
\end{align}
Image inputs and autoregressive sampling for the world model and policy are omitted for clarity.
$L$ indexes the high-level planning horizon.
Optimization proceeds iteratively: at each iteration, $N$ action samples are drawn, simulated, and assigned a cost based on the perceptual distance to the goal at the final predicted frame.
The $M$ samples with the lowest cost are used to update the prior distribution for the next iteration of the algorithm. 
\section{Experiments}
We present our quantitative and qualitative results for the policy conditioned on high-level actions and the Lifted World Model.
Our experiments show the effectiveness of conditioning on the high-level action space compared to goal observations.
We also demonstrate the improved performance and efficiency when planning in high-level action space using the LWM in comparison to PEVA.

\vspace{-0.3cm}
\subsubsection{Base World Model and Nymeria Dataset.}
We use PEVA~\cite{bai2025wholebodypeva} trained on the Nymeria~\cite{ma2024nymeria} dataset as a low-level world model.
Nymeria consists of first-person videos recorded using Project Aria~\cite{engel2023project} glasses and XSens~\cite{roetenberg2009xsens} motion capture suits in continuous 15-minute sessions in 50 indoor and outdoor environments.
The XSens suits provide 3D body motion and joint angle data that are used as the low-level, high-dimensional action space. 
Compared to other datasets, Nymeria presents a combination of a flexible human-like embodiment and challenging hybrid navigation + interaction tasks.

\vspace{-0.3cm}
\subsubsection{Metrics.}
We evaluate the policy and planning tasks by how closely the actions bring the initial pose $p_t$ to the goal pose $p_g$.  
The distance between the final predicted pose $\hat p_g$ and the ground-truth $p_g$ is measured by the mean joint error (MJE)~\cite{ionescu2014human36m} in meters of the \textit{leaf} joints (pelvis, head, and hands), the \textit{intermediate} joints between them, and \textit{all} joints together. 
This provides a ground-truth physical metric for both policy and world model planning performance.

To compute MJE we integrate the predicted actions starting from the initial pose $p_t$ to obtain the predicted final pose $\hat{p}_g$:
\begin{align}
  x_{t+1,\text{pelvis}} &= x_{t,\text{pelvis}} + R(\phi_{t,\text{pelvis}})\delta x_{t,\text{pelvis}} \\
  \phi_{t+1,(\cdot)} &= R^{-1}\big(R(\phi_{t,(\cdot)})R(\delta\phi_{t,(\cdot)})\big)
\end{align}
A forward kinematics procedure (Appendix~\ref{appendix:forward_kinematics}) computes 3D positions for every joint $\{x_\text{pelvis}, \dots, x_\text{left\_hand}\} = \text{forward\_kinematics}(p)$ for both $p_g$ and $\hat p_g$.
The per-joint distances are then computed and averaged over joints for MJE:
\begin{align}
    \text{MJE} = \frac{1}{|\text{joints}|}\sum_{j \in \text{joints}}||\hat{x}_j - x_j||_2.
\end{align}

\vspace{-0.3cm}
\subsubsection{Implementation Details.}
The policy is trained for 10 epochs with a learning rate of $5\times10^{-4}$ using AdamW and a batch size of 256. 
The action sequence length is $T=8$ and the context length is $K_\pi=3$. 
We increase the denoising diffusion network UNet dimensions from NoMaD's $[64, 128, 256]$ to $[256, 384, 512]$ for all experiments. 
Nymeria data preparation matches PEVA~\cite{bai2025wholebodypeva}; sampling is at 4 Hz.
Our experiments use the upper-body PEVA checkpoint with $15$ joints and $48$ action dimensions. World model planning uses up to $6$ CEM iterations and $64$ samples per iteration. 
The world model uses context length $K=8$ and $64$ denoising iterations. 
The action prior is a Gaussian $\mathcal{N}(\bm \mu, \sigma^2I)$ where $\mu=0$ and $\sigma=0.05$ for $a^\text{LL}$ and $\mu=0.5$ and $\sigma=0.3$ for $a^\text{HL}$ which were found by a hyperparameter sweep. 
Waypoints are expressed in normalized image coordinates with boundaries $[0, 1]$. 
The high-level planning horizon is fixed to $L=1$.
The cost function is DreamSIM~\cite{fu2023dreamsim}.

\vspace{-0.3cm}
\subsection{Goal-conditioned policy}
\label{sec:goal_conditioned_policy}
Here we assess the actions generated using our policy trained on Nymeria.
We ablate each addition made to a NoMaD-like diffusion policy in the quantitative results and report \textit{leaf}, \textit{intermediate}, and \textit{all}-joint MJE on the validation set. 
We also include additional experiments on waypoint visibility and on using depth to specify an exact 3D goal.
Qualitative results show that the policy generalizes well to counterfactual waypoints that differ from the waypoints in the data.
In addition, the policy can use image context to infer sensible actions from the same waypoints.
Finally, we show that goal pose $p_g$ cannot be identified from observation $o_g$, limiting $o_g$'s usefulness as a goal for egocentric, embodied agents.

\vspace{-0.3cm}
\subsubsection{Quantitative Results.}
\begin{table*}[t]
\centering
\small
\caption{
Policy Ablations. We evaluate each change applied to the Base Policy. Results are reported on the validation set. 
Mean joint error (MJE) is reported in meters. 
}
\label{tab:policy_performance}
\resizebox{\linewidth}{!}{
\setlength{\tabcolsep}{6pt}
\begin{tabular}{lcccccc}
\toprule
 & \multicolumn{3}{c}{\textbf{Unconditioned MJE}} & \multicolumn{3}{c}{\textbf{Goal-Conditioned MJE}} \\
\cmidrule(lr){2-4} \cmidrule(lr){5-7}
\textbf{Model} &
\textbf{Leaf $\downarrow$} & \textbf{Int. $\downarrow$} & \textbf{All $\downarrow$} &
\textbf{Leaf $\downarrow$} & \textbf{Int. $\downarrow$} & \textbf{All $\downarrow$} \\
\midrule
Initial Distance & 0.445 & 0.419 & 0.426 & 0.445 & 0.419 & 0.426 \\
\midrule
Random Weights & 0.749 & 0.707 & 0.718 & 0.735 & 0.692 & 0.703\\
Base Policy & 0.427 & 0.397 & 0.405 & 0.414 & 0.384 & 0.392 \\
+ architecture changes & 0.406 & 0.375 & 0.384 & 0.388 & 0.359 & 0.367 \\
+ pose context & 0.359 & 0.329 & 0.337 & 0.343 & 0.316 & 0.323 \\
+ waypoint conditioning & \textbf{0.353} & \textbf{0.323} & \textbf{0.331} & 0.262 & 0.236 & 0.243 \\
+ no masking & 0.437 & 0.408 & 0.415 & \textbf{0.243} & \textbf{0.219} & \textbf{0.226} \\
\midrule
+ 3d conditioning & 0.360 & 0.330 & 0.338 & 0.243 & 0.219 & 0.226 \\
+ no masking & 0.415 & 0.391 & 0.398 & 0.223 & 0.202 & 0.208 \\
\bottomrule
\end{tabular}
}
\end{table*}
We report policy performance in Table~\ref{tab:policy_performance}. 
``Unconditioned'' uses no goal information; ``goal-conditioned'' is when either $o_g$ or $a^\text{HL}$ is used to condition the policy.
``Initial distance'' is the MJE between starting pose $p_t$ and goal pose $p_g$ while ``random weights'' refers to actions sampled from an untrained policy.
While the NoMaD baseline outperforms an untrained network, it reduces MJE by only $2.1\text{cm}$ from the initial distance and adding $o_g$ as goal-conditioning reduces it by only another $1.3\text{cm}$.
Adding architectural changes and pose context $\bm p_t$ improve overall performance, but not goal conditioning.
The policy learns coherent action sequences, but the goal observation $o_g$ does not meaningfully steer them toward the goal.

Adding waypoint conditioning improves goal-conditioned MJE by $8.8\text{cm}$ over unconditioned generation.
This improvement indicates that conditioning on $a^\text{HL}$ shifts the low-level action distribution towards actions that better reach goal pose $p_g$.
Furthermore, removing masking improves goal-conditioned performance at the cost of unconditioned MJE.
This tradeoff is acceptable since we prioritize goal-conditioned action generation to lift the world model in later experiments.
Overall, we find that waypoints are an effective and interpretable high-level action space for goal-conditioned generation.

\vspace{-0.3cm}
\subsubsection{Qualitative Action Generations.}
\label{sec:policy_qualitative}
\begin{figure*}[t]
    \centering
    \includegraphics[width=\linewidth]{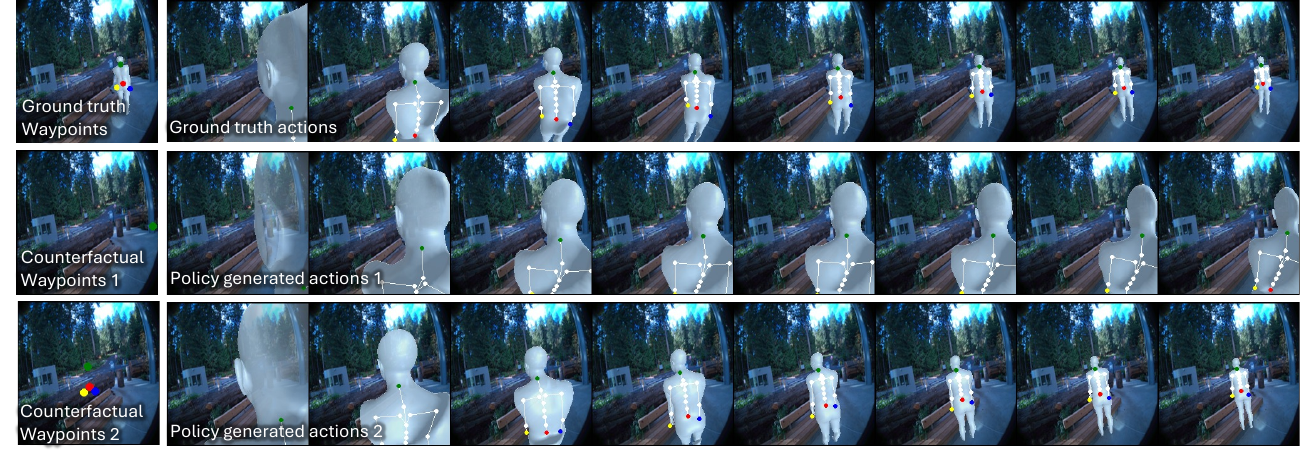}
    \caption{
    Generating actions from waypoints not seen in data.
    \textbf{Ground truth:} the agent walks down the path.
    \textbf{Ex 1:} head \protect\tikz\fill[head] (0,0) circle (0.6ex); waypoint on the right $\rightarrow$ agent moves right, faces left.
    \textbf{Ex 2:} four waypoints above the bench $\rightarrow$ agent climbs onto the bench.
    }
    \label{fig:counterfactual}
\end{figure*}
\begin{figure*}[t]
    \centering
    \includegraphics[width=\linewidth]{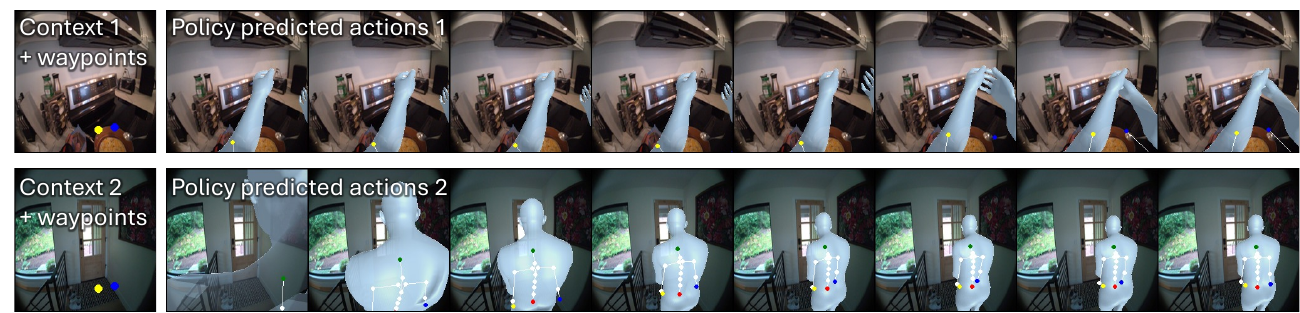}
    \caption{
    Contextually aware action generation.
    The policy predicts different actions from the same waypoints depending on the scene.
    \textbf{Context 1:} agent grasps the pot. \textbf{Context 2:} agent walks right and turns to the left.
    }
    \label{fig:goal_reuse}
\end{figure*}
We visualize generated actions in the ``Predicted actions'' rows of Figure~\ref{fig:lwm_visualizations}. 
Each action in the sequence is shown by projecting the resulting 3D pose onto $o_t$.
The policy controls the agent to both navigate the environment and interact using its hands.
In example 2, it reaches to grasp the doorknob, and in example 3, it raises its hand while holding a camera.

In Figure~\ref{fig:counterfactual}, we evaluate the policy on counterfactual waypoints that do not match the ground truth. 
In counterfactual 1, the policy is given a sparse high-level action of only a head \tikz\fill[head] (0,0) circle (0.6ex); waypoint.
Using this waypoint, the policy moves the agent to the right while facing left, inferring reasonable positions for the unconditioned joints.
In Figure~\ref{fig:goal_reuse} we evaluate the influence of scene context on the generated actions. 
Contexts 1 and 2 use the same waypoints yet result in different action sequences.
We see that in context 1---in front of a stove---the left- \tikz\fill[lhand] (0,0) circle (0.6ex); and right-hand \tikz\fill[rhand] (0,0) circle (0.6ex); waypoints prompt the agent to grasp the pot, while in context 2 the agent walks forward in the open room.

\vspace{-0.3cm}
\subsubsection{Waypoint visibility.}

\begin{table}[t]
\centering
\small
\caption{Goal-conditioned action generation with metrics separated based on the visibility of each joint in the current observation $o_t$. MJE in meters.
}
\label{tab:policy_visibility}
\begin{tabular}{l@{\hspace{1em}}c@{\hspace{1em}}c@{\hspace{1em}}c@{\hspace{1em}}c@{\hspace{1em}}c@{\hspace{1em}}c@{\hspace{1em}}c}
\toprule
 & \multicolumn{3}{c}{\textbf{Visible MJE}} & \multicolumn{3}{c}{\textbf{Not Visible MJE}} \\
\cmidrule(lr){2-4} \cmidrule(lr){5-7}
\textbf{Model} &
\textbf{Leaf $\downarrow$} & \textbf{Int. $\downarrow$} & \textbf{All $\downarrow$} &
\textbf{Leaf $\downarrow$} & \textbf{Int. $\downarrow$} & \textbf{All $\downarrow$} \\
\midrule
Base Policy                & 0.360 & 0.297 & 0.314 & 0.605 & 0.705 & 0.678 \\
+ architecture changes     & 0.340 & 0.278 & 0.295 & 0.566 & 0.659 & 0.634 \\
+ pose context             & 0.305 & 0.254 & 0.268 & 0.483 & 0.551 & 0.533 \\
+ waypoint conditioning    & \textbf{0.260} & \textbf{0.207} & \textbf{0.222} & 0.327 & 0.358 & 0.349 \\
+ no masking         & \textbf{0.251} & \textbf{0.200} & \textbf{0.213} & \textbf{0.284} & \textbf{0.307} & \textbf{0.301} \\
\bottomrule
\end{tabular}
\end{table} 
Since waypoints are defined in observation $o_t$, we investigate whether performance decreases when the joint is not visible. 
Table~\ref{tab:policy_visibility} separates goal-conditioned MJE by the visibility of the joint in $o_t$.
We see a substantial $26.5$cm increase in MJE for observation-conditioned policies when the joint is not visible in the current frame, compared to a modest $12.7$cm increase for waypoint conditioning.
This suggests that a waypoint-conditioned policy learns better whole-body motion patterns that generalize even when joints are not visible at test time.
Furthermore, removing masking improves visible performance while reducing the not-visible performance drop to only $8.8$cm.
This shows that learning with visible waypoints improves performance when they are not in-frame.

\begin{table}[t]
\small
\centering
\caption{Decomposing goal-conditioned action generation into motion generation given a goal pose and predicting the goal pose given the goal observation. MJE in meters.
}
\label{tab:motion_generation_vs_pose_prediction}
  \begin{tabular}{l@{\hspace{1em}}c@{\hspace{1em}}c@{\hspace{1em}}c}
    \toprule
    \textbf{Task} & \textbf{Leaf MJE $\downarrow$} & \textbf{Int. MJE $\downarrow$} & \textbf{All MJE $\downarrow$} \\
    \midrule
    Motion generation & 0.115 & 0.101 & 0.105 \\
    Goal pose prediction & 0.299 & 0.272 & 0.279 \\
    \bottomrule
  \end{tabular}
\end{table}

\vspace{-0.3cm}
\subsubsection{3D Waypoints.}
\label{sec:3d_waypoint_policy}
We assess the efficacy of 2D waypoints by comparing with a policy trained using waypoints augmented with depth values. 
This uses privileged depth information to specify an exact 3D goal, increasing waypoint dimensionality by $50\%$ ($8$ to $12$ dimensions). 
Results are shown in the last two rows of Table~\ref{tab:policy_performance}, where 3D conditioning improves goal-conditioned \textit{all} MJE by at most $2$cm.
We further assess 3D waypoints for CEM planning in Section~\ref{sec:planning_with_3d_waypoints}.

\vspace{-0.3cm}
\subsubsection{Decomposing Action Generation.}
\label{sec:decomposing_action_generation}
We further explore goal-conditioning for egocentric action generation by decomposing it into two separate tasks: motion generation and goal pose prediction.
For \textit{motion generation}, we train a policy to predict joint action sequences with the goal pose $p_g$ as input.
This is a privileged setting where the policy is given the goal pose and must learn to predict the actions to reach it.
For \textit{goal pose prediction}, we replace the denoising UNet with an MLP to directly predict the goal pose $p_g$ conditioned on the goal observation $o_g$. 
See Appendix~\ref{appendix:policy_architecture} for policy architectures.

Both models are evaluated using MJE between the predicted final pose $\hat p_g$ and the ground-truth goal pose $p_g$.
Table~\ref{tab:motion_generation_vs_pose_prediction} shows that predicting an action sequence given the goal pose is easier than predicting the goal pose $p_g$ from $o_g$.
This indicates that egocentric observations are insufficient to identify the goal pose $p_g$ since the embodiment is seldom visible in observation $o_g$ --- the hands may be visible, but the rest of the body is rarely seen.

\subsection{Lifted World Model Planning}
\label{sec:search-based_lwm_planning}
Using the Lifted World Model, we perform planning for hybrid navigation + interaction tasks sampled from the Nymeria dataset.
Each task consists of observations $\bm o_t$, poses $\bm p_t$, and a goal observation $o_g$.
Planning outputs a sequence of actions, and performance is measured by the MJE between the true goal pose $p_g$ and the final pose reached by the planning output.
Note that waypoints are not provided, and the task is to search for waypoints that reach the goal pose.

\vspace{-0.3cm}
\subsubsection{Task Generation.}
We randomly sample start and goal observation pairs from the Nymeria validation set to generate tasks. 
Tasks are constrained to have $\geq1$ joint from the goal pose $p_g$ visible in observation $o_t$ to avoid goals in unseen areas and reduce world model hallucinations.
We also filter tasks for \textit{leaf} MJE $\geq 0.1\text{m}$ between $p_t$ and $p_g$ to avoid trivial stationary tasks.

\vspace{-0.3cm}
\subsubsection{Quantitative Results.}
\begin{table}[t]
\small
\centering
\caption{
Search-based planning results using Cross-Entropy Method on 128 tasks. 6 CEM iterations, 64 samples/iteration.
Mean joint error shown in meters (m). 
}
\label{tab:planning_quantitative_results}
\begin{tabular}{lccc}
\toprule
\textbf{Method} & \textbf{Leaf MJE $\downarrow$} & \textbf{Int. MJE $\downarrow$} & \textbf{All MJE $\downarrow$} \\
\midrule
Initial Distance       & 0.724          & 0.697          & 0.704          \\
Copy Baseline (1-NN)   & 0.696          & 0.663          & 0.672          \\
\midrule
Uncond. Policy         & 0.677          & 0.641          & 0.650          \\
NoMaD                  & 0.605          & 0.578          & 0.585          \\
HDP                    & 0.540   & 0.502          & 0.512          \\
\midrule
PEVA CEM               & 0.637          & 0.608          & 0.616          \\
Lifted CEM (3D)        & 0.453          & 0.407          & 0.420          \\
Lifted CEM (2D, ours)      & \textbf{0.411} & \textbf{0.359} & \textbf{0.374} \\
\bottomrule
\end{tabular}
\end{table}

We report planning results in Table~\ref{tab:planning_quantitative_results}. 
As before, initial distance represents the MJE between $p_t$ and $p_g$.
We also include a 1-NN copy baseline, selecting the action based on nearest $[p_t ;p_g]$.
PEVA CEM searches directly in low-level joint-action space, while Lifted CEM searches for waypoints that generate actions to reach the goal. 
PEVA CEM reduces all MJE by only $8.8\text{cm}$, while Lifted CEM reduces it by $33\text{cm}$.
For each set of waypoints found by Lifted CEM, we sample $64$ joint action sequences and report the average MJE values.
To compare with a policy-only approach, we also evaluate the unconditioned policy, an image-conditioned NoMaD~\cite{sridhar2024nomad} policy, and a Hierarchical Diffusion Policy (HDP)~\cite{ma2024hierarchical}. 
Lifted CEM outperforms all policy-based methods, even ones trained on this goal-observation conditioned planning task.
For ablations of the waypoint joints see Appendix~\ref{appendix:waypoint_joint_ablation}, and for experiments on dexterous manipulation using DexWM~\cite{goswami2025worlddextrous} see Appendix~\ref{appendix:dexterous_manipulation}.

\vspace{-0.3cm}
\subsubsection{Qualitative Planning Visualizations.}
\begin{figure}[t]
  \centering
  \includegraphics[width=0.98\linewidth]{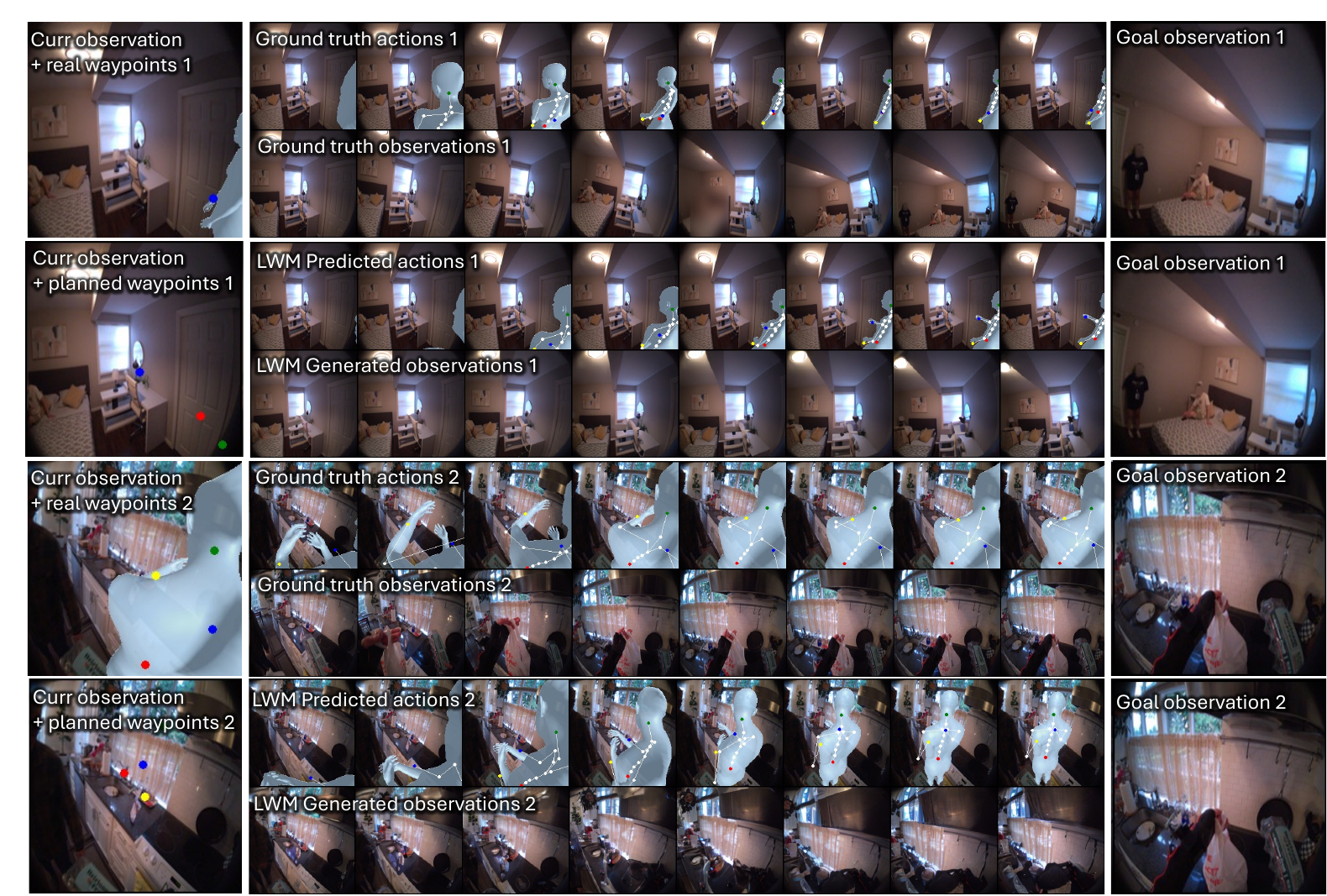}
  \caption{%
    Visualizations of planning solutions. 
    Each row shows from left$\rightarrow$right: the current observation $o_t$ with waypoints, visualized actions (upper row), the resulting observations (lower row), and the goal observation $o_g$.
    \textbf{Task 1:} navigating around the room. \textbf{Task 2:} raising hands to set down the plastic bag on the counter.
    }
  \label{fig:planning_vis}
\end{figure}

We visualize action and observation sequences using waypoints from CEM planning in Figure~\ref{fig:planning_vis}.
Planning with Lifted CEM effectively reproduces the ground truth actions.
In task 1, the agent moves to the right following the pelvis \tikz\fill[pelvis] (0,0) circle (0.6ex); waypoint and faces left, while in task 2, it moves its hands over the counter to set down the plastic bag.
We find that there are key waypoints that drive the action: 
(right hand \tikz\fill[rhand] (0,0) circle (0.6ex);, pelvis \tikz\fill[pelvis] (0,0) circle (0.6ex);) 
in task 1 and 
(left hand \tikz\fill[lhand] (0,0) circle (0.6ex);, right hand \tikz\fill[rhand] (0,0) circle (0.6ex);) 
in task 2.
Similar to how the policy can handle missing waypoints (Section~\ref{sec:policy_qualitative}), it can ignore outlier waypoints found by search: head \tikz\fill[head] (0,0) circle (0.6ex); in task 1 and pelvis \tikz\fill[pelvis] (0,0) circle (0.6ex); in task 2. 
The policy in Lifted CEM is trained to produce realistic movements, while PEVA CEM may produce unnatural joint angles during planning. 
See Appendix~\ref{appendix:peva_planning_vis} for PEVA planning visualizations.

\vspace{-0.3cm}
\subsubsection{Planning Efficiency.}
\begin{figure}[t]
  \centering
  \includegraphics[width=0.7\linewidth]{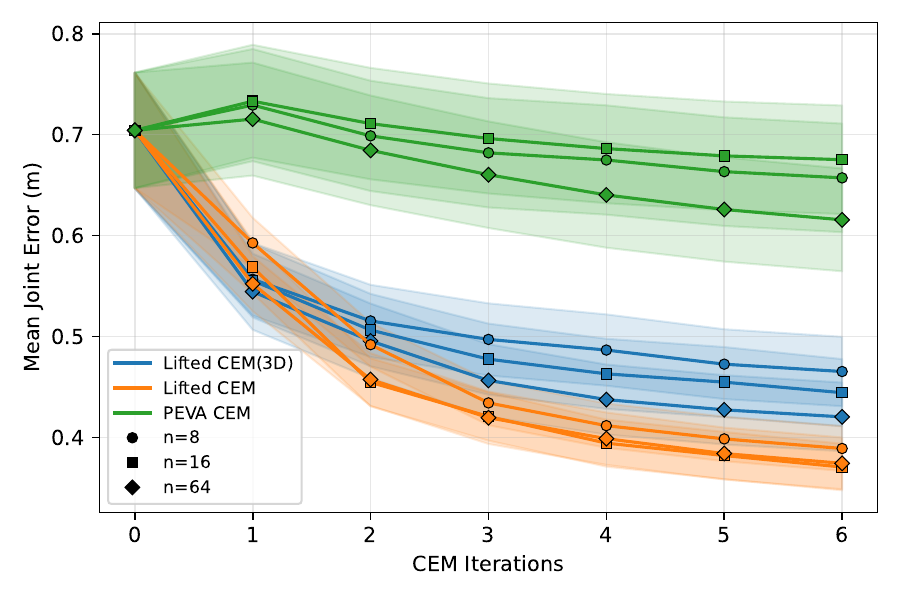}
  \caption{
    Planning performance for different CEM budgets. Searching in our lifted action space scales better than in joint space. 
    3D waypoints are more difficult to search.
    }
  \label{fig:planning_cem_parameters_wrt_metric}
\end{figure}
Planning using CEM with a large world model can be computationally expensive.
Thus, we compare the performance of searching over high-level (Lifted CEM) vs. low-level actions (PEVA CEM) for different levels of compute.
We measure the MJE of both methods for varying CEM iterations and samples per iteration.
For compute measurements see Appendix~\ref{appendix:compute}.

Results are shown in Figure~\ref{fig:planning_cem_parameters_wrt_metric}.
We show the cumulative min over CEM iterations, see Appendix~\ref{appendix:no_cumulativemin} for without.
Planning in high-level space outperforms planning in low-level joint action space for all numbers of samples and CEM iterations.
Furthermore, we find that PEVA CEM worsens MJE after one step, which highlights the difficulty of naive search. 
In summary, lifting the abstraction of the world model yields better solutions with less compute.

\vspace{-0.3cm}
\subsubsection{Lifting Using 3D Waypoints.}
\label{sec:planning_with_3d_waypoints}
We evaluate a world model lifted to use 3D waypoints in search-based planning.
The results are plotted alongside Lifted CEM in Figure~\ref{fig:planning_cem_parameters_wrt_metric}. 
3D waypoints perform similarly after one CEM step, but lag behind with further iterations due to the larger action dimensionality ($12$ vs. $8$) and the difficulty of finding a correct depth value.

\vspace{-0.3cm}
\subsubsection{Varying Goal Horizons.}
\begin{figure}[t]
  \centering
  \begin{minipage}{0.49\linewidth}
    \centering
    \includegraphics[width=\linewidth]{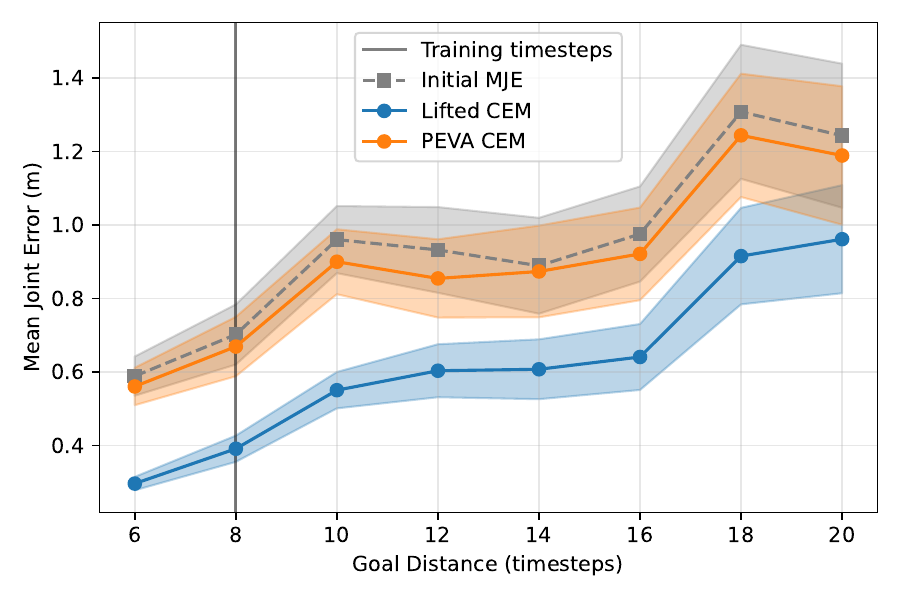}
    \caption{Planning results for varying goal horizons. The vertical line shows the policy training distribution.}
    \label{fig:planning_cem_parameters_wrt_timesteps}
  \end{minipage}
  \begin{minipage}{0.49\linewidth}
    \centering
    \includegraphics[width=\linewidth]{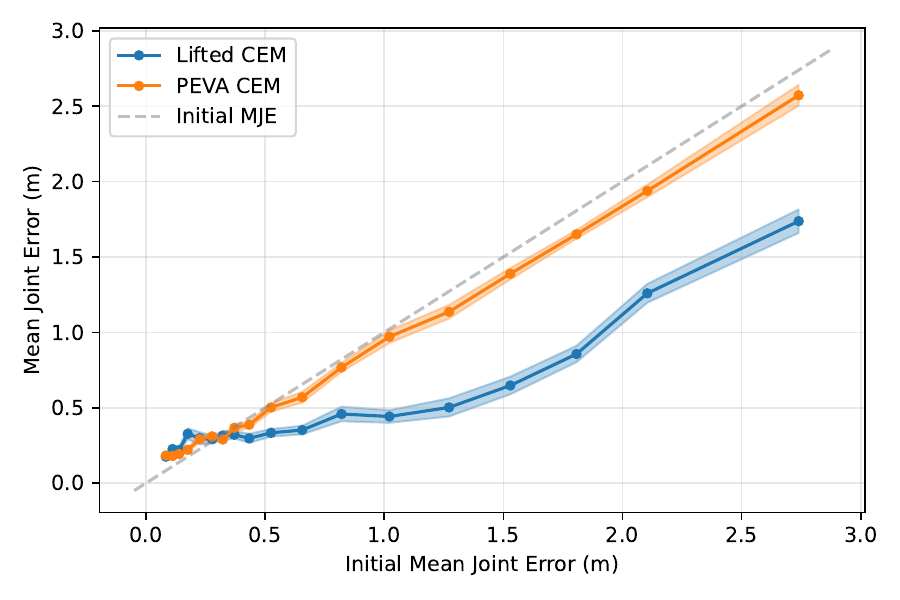}
    \caption{Planning on tasks of varying inital MJE. Tasks are grouped using 20 quantile buckets.}
    \label{fig:planning_cem_parameters_wrt_init_mje}
  \end{minipage}
\end{figure}
We also report planning results for goals sampled from different timesteps into the future and different initial MJE distances. 
Figure~\ref{fig:planning_cem_parameters_wrt_timesteps} shows performance when goals are sampled from varying time horizons.
In this setting, PEVA CEM plans for actions of sequence length equal to the goal horizon whereas Lifted CEM is fixed at $8$ due to fixed policy length. 
Despite this, we find that Lifted CEM outperforms PEVA CEM over all goal horizons.

Figure~\ref{fig:planning_cem_parameters_wrt_init_mje} plots performance against initial MJE, with tasks grouped into 20 quantile buckets. 
The $y=x$ line is a baseline where no action is taken, where the initial and final MJE are equal.
Lifted CEM outperforms PEVA CEM at most initial distances while both methods perform worse when initial MJE $\leq0.3$m.

\vspace{-0.3cm}
\subsubsection{Unseen Environments.}
\label{sec:unseen_environments}
\begin{table}[t]
\centering
\small
\caption{Lifted CEM planning on tasks in unseen environments. MJE in meters.
}
\label{tab:planning_unseen_environments}
\begin{tabular}{lccc}
\toprule
\textbf{Method}        & \textbf{Leaf MJE $\downarrow$} & \textbf{Int. MJE $\downarrow$} & \textbf{All MJE $\downarrow$} \\
\midrule
Initial Distance       & 0.618             & 0.589             & 0.597            \\
PEVA CEM               & 0.572             & 0.546             & 0.553            \\
Lifted CEM (ours)      & 0.373             & 0.318             & 0.333            \\
Lifted CEM (held-out)  & 0.404             & 0.346             & 0.362            \\
\bottomrule
\end{tabular}
\vspace{-1em}
\end{table}
To test the generalization of our lifting methodology, we train a policy on a new training set where videos from Nymeria locations 6, 19, and 34 are replaced with videos from other locations.
Then, we use this policy to lift the same PEVA model and evaluate the lifted model on tasks from held-out locations 6, 19, and 34.

Results are presented in Table~\ref{tab:planning_unseen_environments}.
Lifted CEM using the held-out policy outperforms PEVA CEM and is only slightly worse than using the base policy trained on all environments.
Therefore, the short-term lightweight policy design generalizes well to unseen environments.

\section{Related Work}
\vspace{-0.3cm}
\subsubsection{World Models and Planning.}
Many world model works explore planning: PlaNet~\cite{hafner2019learningplanet}, TD-MPC2~\cite{hansentdmpc2}, DINO-WM~\cite{zhou2025dinowm} and VJEPA-2~\cite{assran2025vjepa,mur2026v} plan in latent space, while UniSim~\cite{yang2024unisim} and NWM~\cite{bar2025navigationnwm} are used downstream to plan in pixel space.
Among these works, cross-entropy method~\cite{rubinstein199789crossentropymethodcem} is the \textit{de facto} planning approach, while MPPI~\cite{williams2017model} also sees use. 
DINO-WM~\cite{zhou2025dinowm} and Parthasarathy et al.~\cite{parthasarathy2025closinggradientplanning} explore using gradient-based methods. 
DynaGuide~\cite{du2025dynaguidesteeringdiffusionpolices} guides a diffusion policy via gradients through a latent world model.
The Dreamer~\cite{hafnerdreamerv1, hafnermasteringdreamerv2, hafner2025dreamerv3} series learns a world model for policy training.
PEVA~\cite{bai2025wholebodypeva} trains a whole-body conditioned world model, and DexWM~\cite{goswami2025worlddextrous} conditions on hands.
To our knowledge, our work is the first to use a world model to plan for complex, egocentric, human-like embodiments, and the first to lift a low-level world model to a higher level of abstraction to make this challenging task tractable.

\vspace{-0.3cm}
\subsubsection{Embodied, Egocentric, and Hierarchical Policies.}
NoMaD~\cite{sridhar2024nomad} tackles navigation from an egocentric point of view with a point-robot embodiment while Diffusion Policy~\cite{chi2025diffusionpolicy} uses a similar architecture for manipulation with articulated robots. 
Nachum et al.~\cite{nachum2018datahrl} and later HIQL~\cite{park2023hiql} are works that learn hierarchical policies. 
iDP3~\cite{ze2025generalizable} controls a humanoid at a joint level from an egocentric 3D camera.
Large-scale models include GR00T-N1~\cite{bjorck2025gr00t}, which controls an articulated humanoid, and VLA approaches like OpenVLA~\cite{kim25openvla} and Octo~\cite{2024octo}, which tend to focus on robotic arms and exocentric manipulation tasks.
No prior works address navigation and manipulation for embodiments from an egocentric view, which is a key affordance of human-like agents, nor do they consider the use of a low-dimensional, finite, and searchable input space for goal conditioning.

\vspace{-0.3cm}
\subsubsection{Motion Generation.}
Motion generation aims to learn a realistic distribution of human motion controlled by text, action labels, or keypoint constraints.
MDM~\cite{tevethumanmdm} and MotionDiffuse~\cite{zhang2024motiondiffuse} use diffusion for text-conditioned motion generation and PriorMDM~\cite{shafirhumanpriormdm} extends this with dense end-effector and keypoint conditioning.
OmniControl~\cite{xieomnicontrol} supports sparse 3D keypoint conditioning over arbitrary joints and timesteps.
CLoSD~\cite{tevet2025closd} and MaskedMimic~\cite{tessler2024maskedmimic} both utilize a physics simulator: the former closes the generation loop with simulation, while the latter performs motion inpainting.
GOAL~\cite{taheri2022goal} focuses on hand-object grasping for a target object in an open-loop setting.
All of these works control a human embodiment, but do not address reaching a precise goal state in an environment from egocentric observations.
Our waypoint approach shares some similarity with keypoint conditioning, but specifies sparse 2D joint positions for the final goal pose, relying on the policy to capture short-term motion patterns instead of a dense sequence of 3D joint positions.


\section{Limitations and Future Work}
Currently, our method does not explicitly control head orientation. 
This allows the agent to look in various directions which can affect observations generated by the world model. 
Explicit control of the head pose may be helpful --- specifying an orientation or marking a target to look at are possible approaches. 

Also, waypoints cannot specify goal poses that lie fully outside the current observation. 
This limitation is acceptable as world models often hallucinate when generating unseen spaces.
However, as world models improve in consistency, waypoints that extend beyond the current observation may become an interesting direction.

\section{Conclusion}
We present a method for lifting a low-level world model to a higher level of abstraction by training a goal-conditioned policy that maps high-level actions to sequences of low-level actions, and instantiate it for a human-like embodiment using waypoints as the high-level action space.
We show that embodied, egocentric agents benefit from specifying goals as waypoints grounded in the present observation, and that search-based planning with the resulting Lifted World Model is $3.8\times$ more effective than planning in low-level joint space, at lower compute cost, and generalizes to environments unseen by the policy. 
Notably, our approach is lightweight, scalable, and requires no additional data collection or annotation beyond what is already used to train the world model. 
Moving forward, we believe lifting is a promising direction for keeping world models controllable and plannable as they scale to richer embodiments and more complex tasks; exploring alternative high-level action spaces and embodiments is a natural next step.

\newpage

\bibliography{eccv2026/main}
\bibliographystyle{splncs04}

\newpage
\appendix
\renewcommand{\theHsection}{appendix.\Alph{section}}
\onecolumn
\section{Lifting a Dexterous Manipulation World Model}
\label{appendix:dexterous_manipulation}
We apply our lifting approach to DexWM~\cite{goswami2025worlddextrous}. 
Frames are sampled over $2$ seconds at $2$Hz for a total of $4$ frames.
The high-level actions are defined as the:
\begin{equation}
    a^\text{HL} = \{w_\text{left\_wrist}, w_\text{right\_wrist}, w_\text{left\_index\_tip}, w_\text{right\_index\_tip}\}
\end{equation}
for a total of $8$ dimensions. 
We increase the denoising Unet to $[512, 768, 1024]$ dimensions and $30$ denoising timesteps.

We follow the same procedure for planning with CEM using $\sigma=60$px in observation $o_t \in [0, 1]^{224\times224}$.
We use a latent L2 cost
\begin{equation}
    \text{cost} = ||\hat z_g - z_g||_2
\end{equation}
instead of DreamSIM because DexWM predicts DINOv2~\cite{oquabdinov2} latents. 
Instead of sampling autoregressively using DexWM, we combine the 4 action steps into a single-step action due to rollout instability (see Appendix~\ref{appendix:dexwm_rollout_instability}).
\subsection{Planning results}
We present our preliminary planning results on $256$ tasks sampled from EgoDex~\cite{egodex}.
$8$ CEM iterations were performed with $256$ samples per iteration.
Surprisingly, Base CEM is unable to find any solution in an $142$ dimensional action space.
\begin{table}[h]
\centering
\begin{tabular}{lc}
\toprule
\textbf{Method} & \textbf{All MJE (cm) $\downarrow$} \\
\midrule
Initial Distance       & 10.46          \\
\midrule
Base CEM               & 13.81          \\
Lifted CEM       & \textbf{7.98}  \\
\bottomrule
\end{tabular}
\caption{All-MJE for dexterous manipulation using DexWM}
\label{tab:dexterous_manipulation_planning_results}
\end{table}
\vspace{-4em}
\subsection{Rollout instability}
\label{appendix:dexwm_rollout_instability}
We find that DexWM exhibits rollout instability. 
Sampling autoregressively for $4$ or $8$ steps over $2$ seconds produces a higher L2 loss than summing the action sequence over all timesteps and performing a single prediction.
Over $256$ tasks, the L2 latent loss from predicting the final state using the ground-truth actions is shown in Table~\ref{tab:rollout_instability}.
\begin{table}[h]
\centering
\setlength{\tabcolsep}{18pt}
\begin{tabular}{ccc}
\toprule
1-step & 4-step & 8-step \\
\midrule
0.818 & 0.868 & 0.994 \\
\bottomrule
\end{tabular}
\caption{Goal state L2 error from ground truth actions}
\label{tab:rollout_instability}
\end{table}
\vspace{-3em}
\newpage

\section{Cost-function convergence during world model planning}
\begin{figure}[H]
  \centering
  \includegraphics[width=0.8\linewidth]{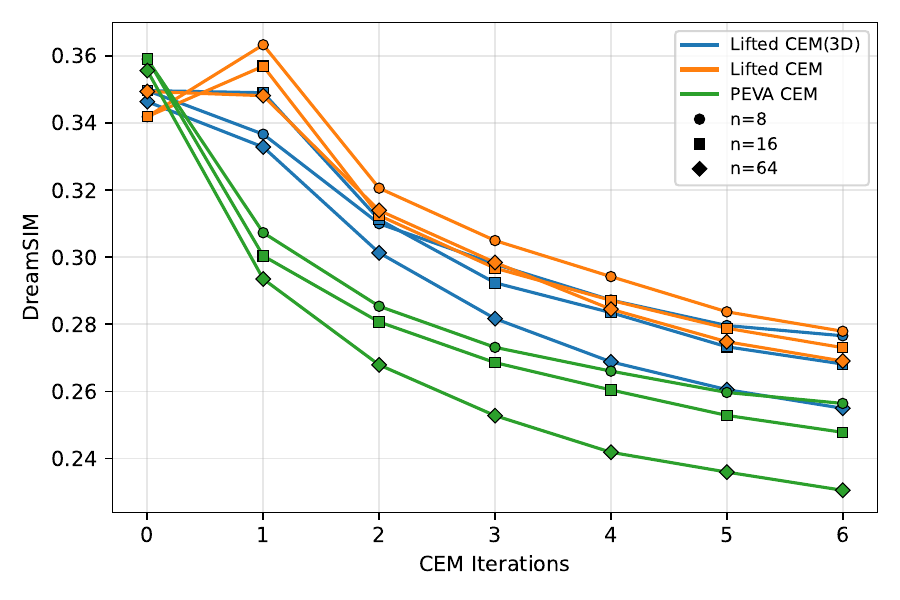}
  \caption{%
    Cost function convergence with respect to CEM iterations and number of samples. 
  }
  \label{fig:planning_cem_parameters_wrt_costfn}
\end{figure}
Across all methods, more samples leads to a lower DreamSIM perceptual distance.
However, Lifted CEM has the highest perceptual distance while PEVA has the lowest perceptual distance.
This suggests a mismatch between our MJE metric and the perceptual distance metric.
This makes sense as the goal observation in general is often not informative of the present pose of the embodiment.
This means the body can be in any manner of incorrect poses despite a good DreamSIM, or the body can be in the correct pose but have a higher perceptual distance due to a slightly altered head pose.

\subsection{Policy as a regularizer}
In general this presents the interesting possibility that a good policy acts as a regularizer for the planning procedure, limiting it to only plausible human actions. 
It may be possible that CEM in low-level joint space on a per-timestep level might be hacking the perceptual similarity objective.

\newpage 
\section{Waypoint Joint Ablation}
\label{appendix:waypoint_joint_ablation}
\vspace{-1em}
We ablate the choice of joints for both goal-conditioned policy training as well as planning. 
See Table~\ref{tab:joint_ablations} for results.
We find that our method is robust to the choice of joints used for high-level waypoint actions.
Planning was performed with 6 iterations, 8 samples for 64 tasks.
\vspace{-1em}
\begin{table}[h]
\footnotesize
\centering
\setlength{\tabcolsep}{4pt}
\renewcommand{\arraystretch}{0.85}
\begin{tabular}{lcccccc}
\toprule
& \multicolumn{3}{c}{\textbf{Goal-Conditioned}} & \multicolumn{3}{c}{\textbf{Planning}} \\
\cmidrule(lr){2-4} \cmidrule(lr){5-7}
\textbf{Method} & \textbf{Leaf $\downarrow$} & \textbf{Int. $\downarrow$} & \textbf{All $\downarrow$} & \textbf{Leaf $\downarrow$} & \textbf{Int. $\downarrow$} & \textbf{All $\downarrow$} \\
\midrule
Pelvis         & 0.280 & 0.247 & 0.255 & 0.482 & 0.451 & 0.460 {\scriptsize $\pm$0.043} \\
Head           & 0.299 & 0.266 & 0.275 & 0.395 & 0.353 & 0.365 {\scriptsize $\pm$0.039} \\
Hands          & 0.256 & 0.235 & 0.240 & 0.430 & 0.380 & 0.394 {\scriptsize $\pm$0.039} \\
Pelvis+Head    & 0.268 & 0.233 & 0.242 & 0.467 & 0.432 & 0.442 {\scriptsize $\pm$0.034} \\
Pelvis+Hands   & 0.250 & 0.227 & 0.233 & 0.422 & 0.371 & 0.385 {\scriptsize $\pm$0.036} \\
Head+Hands     & 0.247 & 0.224 & 0.230 & 0.406 & 0.357 & 0.371 {\scriptsize $\pm$0.038} \\
\midrule
Pelvis+Head+Hands& 0.243 & 0.219 & 0.226 & 0.431 & 0.394 & 0.405 {\scriptsize $\pm$0.035} \\
\bottomrule
\end{tabular}
\caption{Goal-conditioned waypoints is robust to choice of joints.}
\label{tab:joint_ablations}
\end{table}
\vspace{-1em}

\section{Policy and PEVA Compute Measurements}
See Table~\ref{tab:policy_vs_wm_compute} for model compute differences.
\label{appendix:compute}
\begin{table}[h]
\footnotesize
\centering
\setlength{\tabcolsep}{4pt}
\renewcommand{\arraystretch}{0.85}
\begin{tabular}{lrrr}
\toprule
\textbf{Model} & \textbf{Params (M)} & \textbf{Wall-clock (ms)} & \textbf{FLOPs (GFLOPs)} \\
\midrule
Policy $\pi_\theta$ & 58.1    & 63.9     & 605.4 \\
PEVA                & 1{,}060.1 & 48{,}203.9 & 12{,}438{,}305.4 \\
\bottomrule
\end{tabular}
\caption{Policy vs. World Model Compute}
\label{tab:policy_vs_wm_compute}
\vspace{-3mm}
\end{table}

\section{Generalization of the Lifted World Model}
We evaluate the generalization of our lifting method in subsection~\ref{sec:unseen_environments} by performing planning in environments held out during policy training. 
We find that our policy performs well and samples reasonable actions even in unseen environments.

We note that PEVA has been trained on these environments, but neither model has seen the planning tasks which come from validation videos.
This is important as using a world model to simulate unseen environments amounts to planning by imagination.
Furthermore, our contribution is the proposed lifting procedure which only involves training the lifting policy.
\newpage

\section{Waypoint Masking Experiments} \label{appendix:waypoint_masking}
\vspace{-1em}
We preliminarily explore masking waypoints during training so the policy can better handle sparse inputs at inference.
For example, only a pelvis waypoint can be used to navigate, leaving the other joint targets unspecified.
Masking also teaches the policy to infer the target joint position when a waypoint is missing, such as when it is out of frame.
During training, half the time no waypoints are masked, and half the time 
each waypoint is independently masked with probability 0.5.

In our limited training run, waypoint masking sacrifices some goal-conditioned policy performance and performs better on unconditioned action generation.
However, interestingly, in planning, the waypoint masking policy performs better, particularly on the Leaf joints, by a whole 1.1cm. 
This indicates the possibility that randomly masking waypoints teaches the model to better-infer missing waypoints.
CEM planning often only finds a subset of correctly situated waypoints and the waypoint masked policy is better able to infer the remaining ones.

It is possible that further training and more policy capacity may help close the goal-conditioned gap which we leave for future works.

\vspace{-1em}
\begin{table*}[h]
\centering
\small
\vspace{-0.4em}
\caption{
Waypoint masking policy performance compared to no masking.
}
\resizebox{\linewidth}{!}{
\setlength{\tabcolsep}{6pt}
\begin{tabular}{lcccccc}
\toprule
 & \multicolumn{3}{c}{\textbf{Unconditioned MJE}} & \multicolumn{3}{c}{\textbf{Goal-Conditioned MJE}} \\
\cmidrule(lr){2-4} \cmidrule(lr){5-7}
\textbf{Model} &
\textbf{Leaf $\downarrow$} & \textbf{Int. $\downarrow$} & \textbf{All $\downarrow$} &
\textbf{Leaf $\downarrow$} & \textbf{Int. $\downarrow$} & \textbf{All $\downarrow$} \\
\midrule
Initial Distance & 0.445 & 0.419 & 0.426 & 0.445 & 0.419 & 0.426 \\
\midrule
no masking & 0.415 & 0.391 & 0.398 & 0.223 & 0.202 & 0.208 \\
waypoint masking & 0.364 & 0.335 & 0.343 & 0.263 & 0.236 & 0.243 \\
\bottomrule
\end{tabular}}
\end{table*}
\vspace{-1em}
\begin{table}[H]
\small
\centering
\caption{
Planning results with waypoint masking compared to no masking with 64 samples per task for 64 tasks.
}
\begin{tabular}{lccc}
\toprule
\textbf{Method} & \textbf{Leaf MJE $\downarrow$} & \textbf{Int. MJE $\downarrow$} & \textbf{All MJE $\downarrow$} \\
\midrule
Initial Distance       & 0.724          & 0.697          & 0.704          \\
\midrule
Lifted CEM (2D, ours)      & 0.399 & \textbf{0.349} & \textbf{0.364} \\
Waypoint Masking &  \textbf{0.388} & \textbf{0.349} & \textbf{0.360} \\
\bottomrule
\end{tabular}
\end{table}

\newpage
\section{CEM iteration graphs without cumulative min for each step}
\label{appendix:no_cumulativemin}
\begin{figure}[H]
  \centering
  \includegraphics[width=0.7\linewidth]{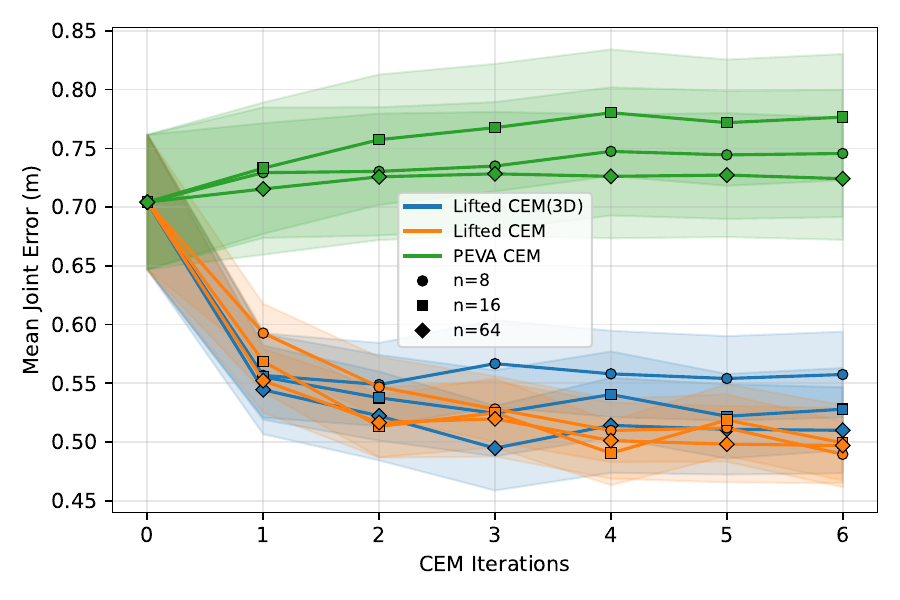}
  \caption{%
    MJE over CEM iterations without cumulative minimum.
  }
  \label{fig:cem_iterations_mje_without_cumulative_min}
\end{figure}
\begin{figure}[H]
  \centering
  \includegraphics[width=0.7\linewidth]{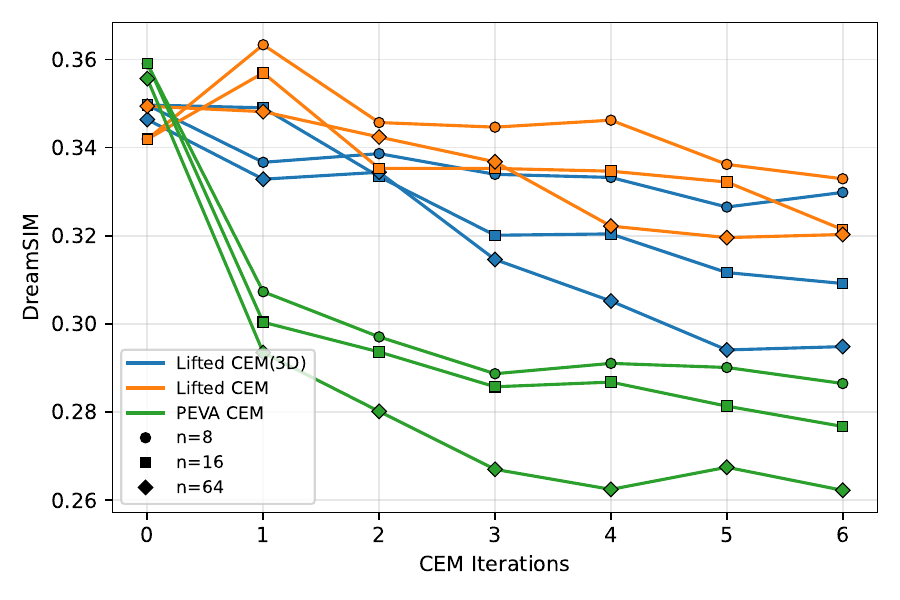}
  \caption{%
    Cost function convergence with respect to CEM iterations and number of samples without cumulative minimum.
  }
  \label{fig:cem_iterations_cost_without_cumulative_min}
\end{figure}
We do not take the cumulative minimum over these steps. 
Lifted CEM is still much more effective than PEVA CEM. 
While noisier, Lifted CEM also outperforms Lifted CEM with a 3D-augmented policy.

\newpage
\section{Forward Kinematics}
\label{appendix:forward_kinematics}
We use a forward kinematics computation based on the XSens~\cite{roetenberg2009xsens} human model to compute joint positions in 3D space. 
The pose representation used in our paper directly specifies the pelvis position in 3D space as the root of our embodiment $ x_\text{pelvis}$. 
The 3D position of a child joint is computed in a world frame. 
\begin{align}
     x_\text{child} = x_\text{parent} + R(\phi_\text{parent})\Delta x_\text{parent,child}
\end{align}
The Euler angle $\phi_\text{parent}$ is the parent orientation in the world frame.
$\Delta x_\text{parent,child}$ is the spatial offset representing the bone defined in the parent frame. 
The individual bone lengths vary for each participant. 

The kinematic parent-child relationships are shown in Table~\ref{tab:parent-child-kinematic-chain}.
\begin{table}[h]
\centering
\begin{tabular}{ll}
\hline
Kinematic parent & Child \\
\hline
N/A & Pelvis \\
Pelvis & L5 \\
L5 & L3 \\
L3 & T12 \\
T12 & T8 \\
T8 & Neck \\
Neck & Head \\
T8 & R\_Shoulder \\
R\_Shoulder & R\_UpperArm \\
R\_UpperArm & R\_Forearm \\
R\_Forearm & R\_Hand \\
T8 & L\_Shoulder \\
L\_Shoulder & L\_UpperArm \\
L\_UpperArm & L\_Forearm \\
L\_Forearm & L\_Hand \\
\hline
\end{tabular}
\caption{Kinematic tree parent-child relationships}
\label{tab:parent-child-kinematic-chain}
\end{table}

\newpage
\section{Policy Architecture}
\label{appendix:policy_architecture}
Below are the policy architectures used in this paper.
We also present the motion generation and goal prediction models used in Section~\ref{sec:decomposing_action_generation}.
\begin{figure}[h]
    \centering
    \includegraphics[width=0.85\linewidth]{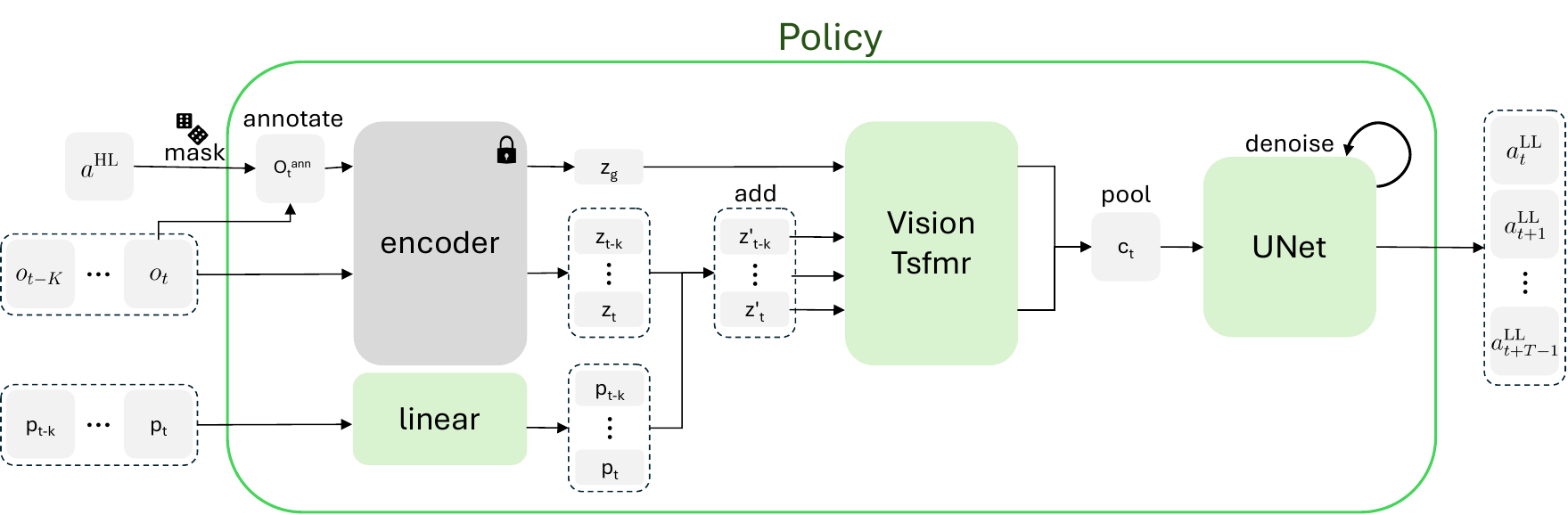}
    \caption{Policy architecture}
    \label{fig:policy_architecture}

    \vspace{1em}

    \includegraphics[width=0.85\linewidth]{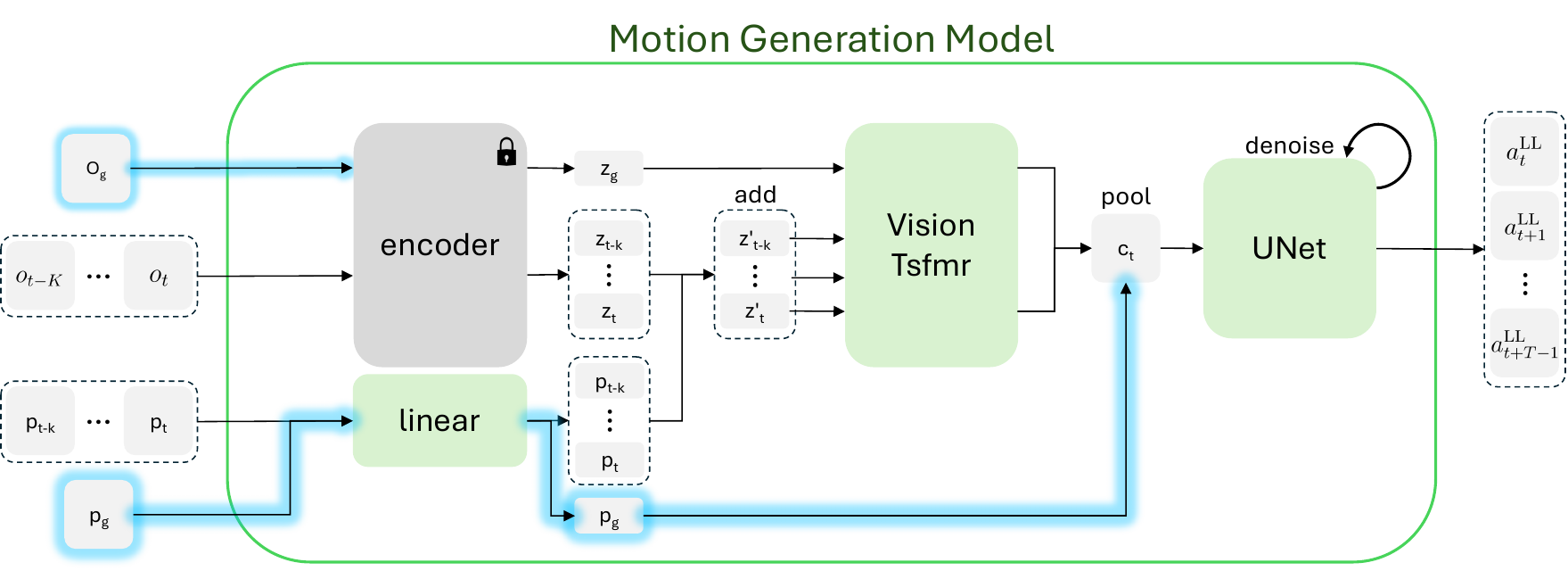}
    \caption{Motion Generation model architecture; changes highlighted in blue.}
    \label{fig:motion_generation_architecture}

    \vspace{1em}

    \includegraphics[width=0.85\linewidth]{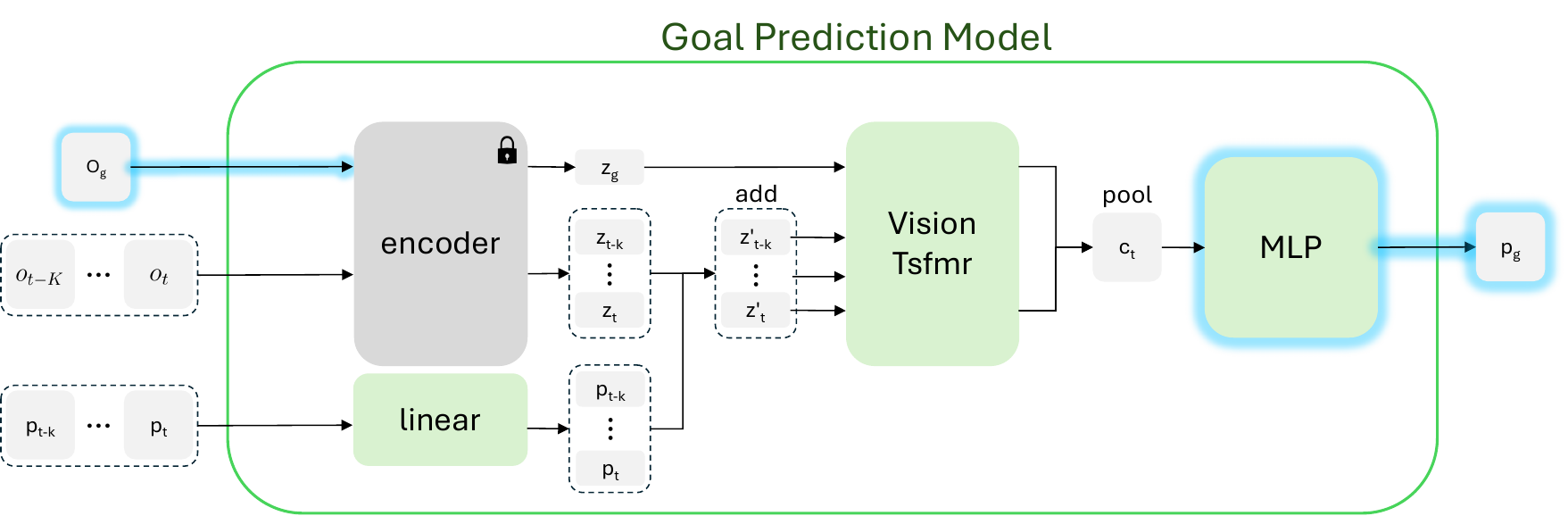}
    \caption{Goal prediction model architecture; changes highlighted in blue.}
    \label{fig:goal_prediction_architecture}
\end{figure}

\newpage
\newpage

\section{Additional Planning Visualizations}
Additional planning visualizations.
\begin{figure}[h]
  \centering
  \includegraphics[width=0.98\linewidth]{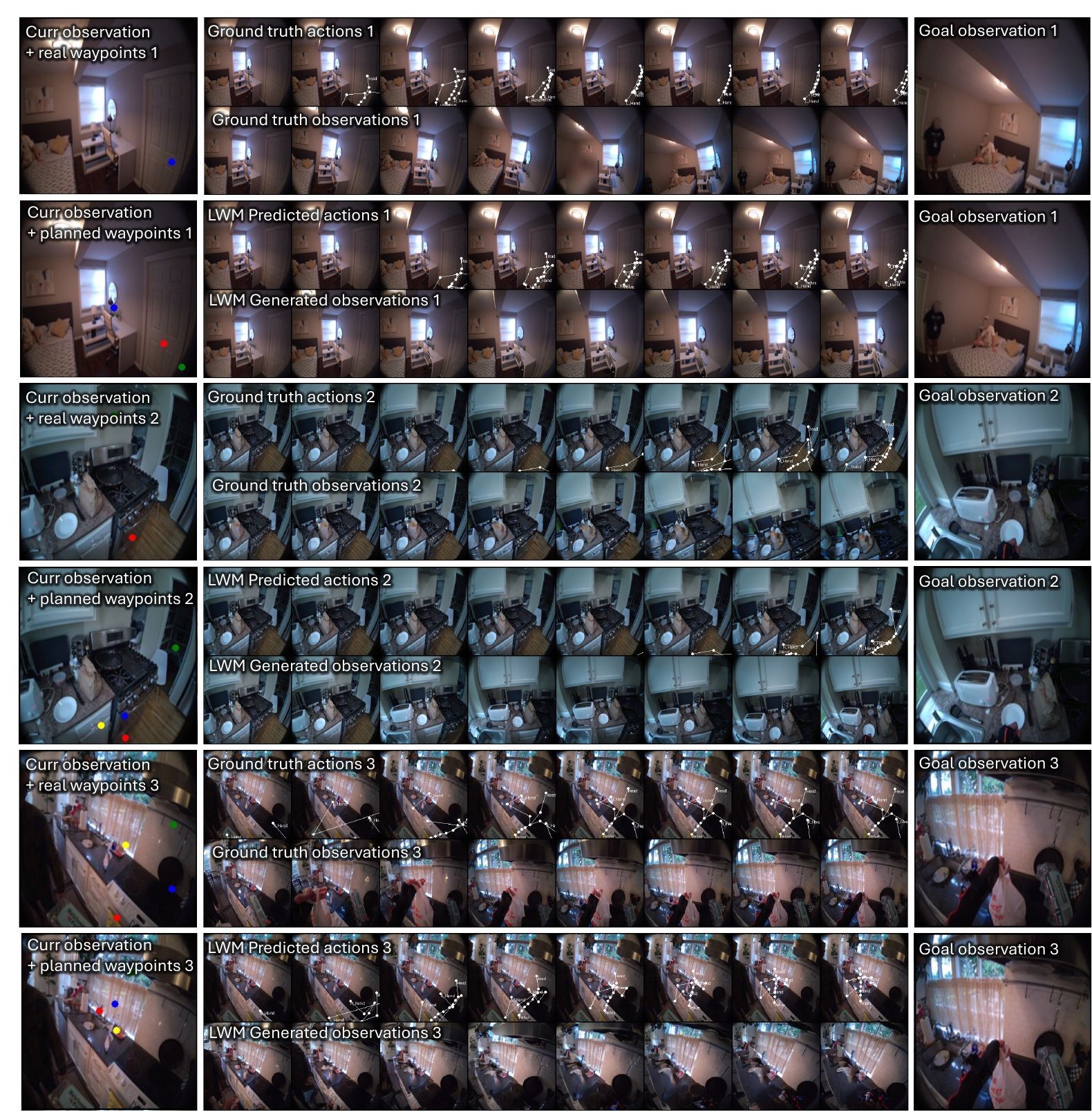}
  \caption{%
    Waypoint CEM with additional tasks, and no SMPL mesh.
    }
  \label{fig:extra_planning_vis_skeleton}
\end{figure}
\newpage
\begin{figure}[h]
  \centering
  \includegraphics[width=0.98\linewidth]{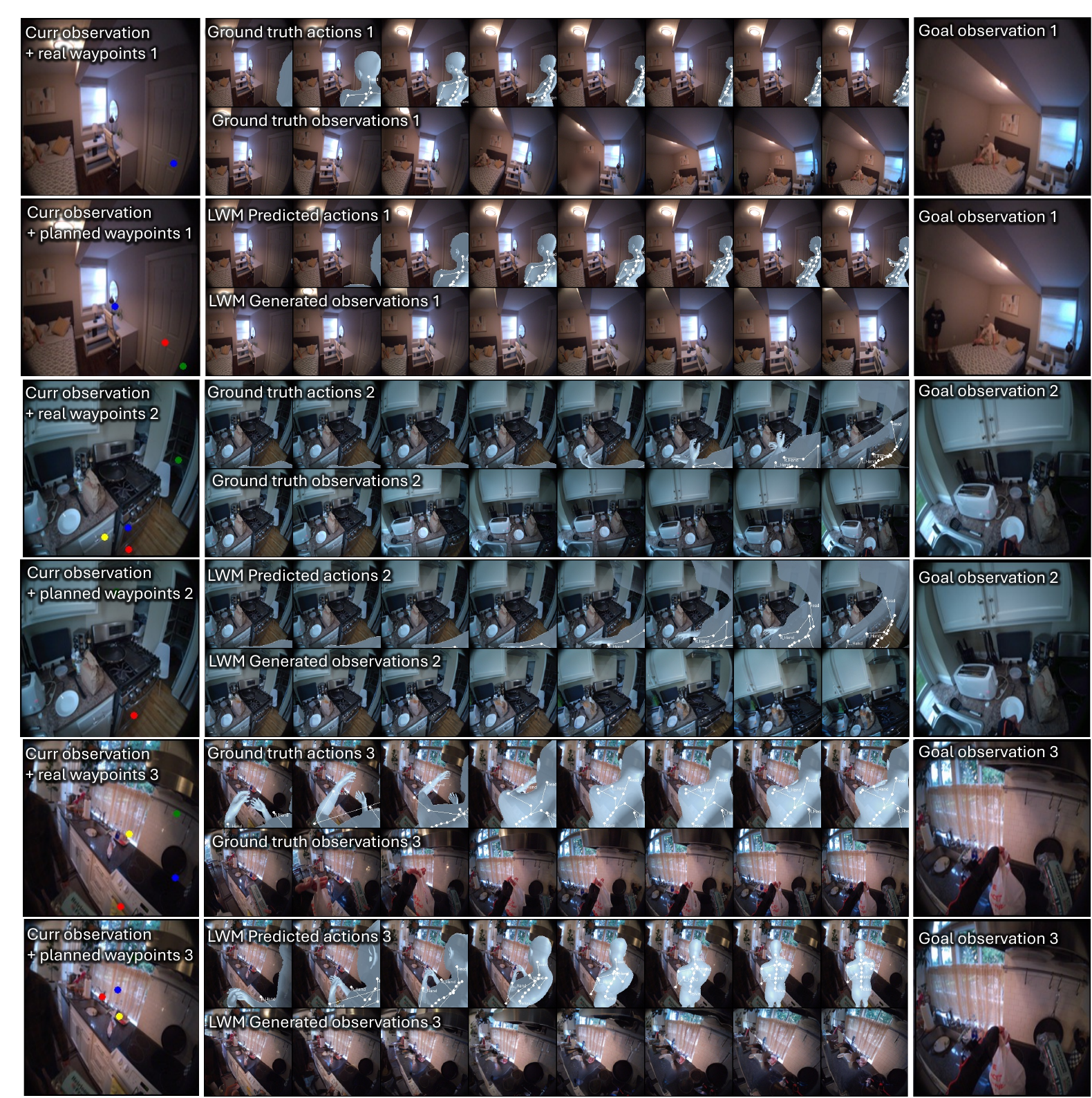}
  \caption{%
    Waypoint CEM with additional tasks, with mesh.
    }
  \label{fig:extra_planning_vis_skinned}
\end{figure}

\newpage
\section{PEVA Planning Visualizations}
\label{appendix:peva_planning_vis}
\begin{figure}[h]
  \centering
  \includegraphics[width=0.98\linewidth]{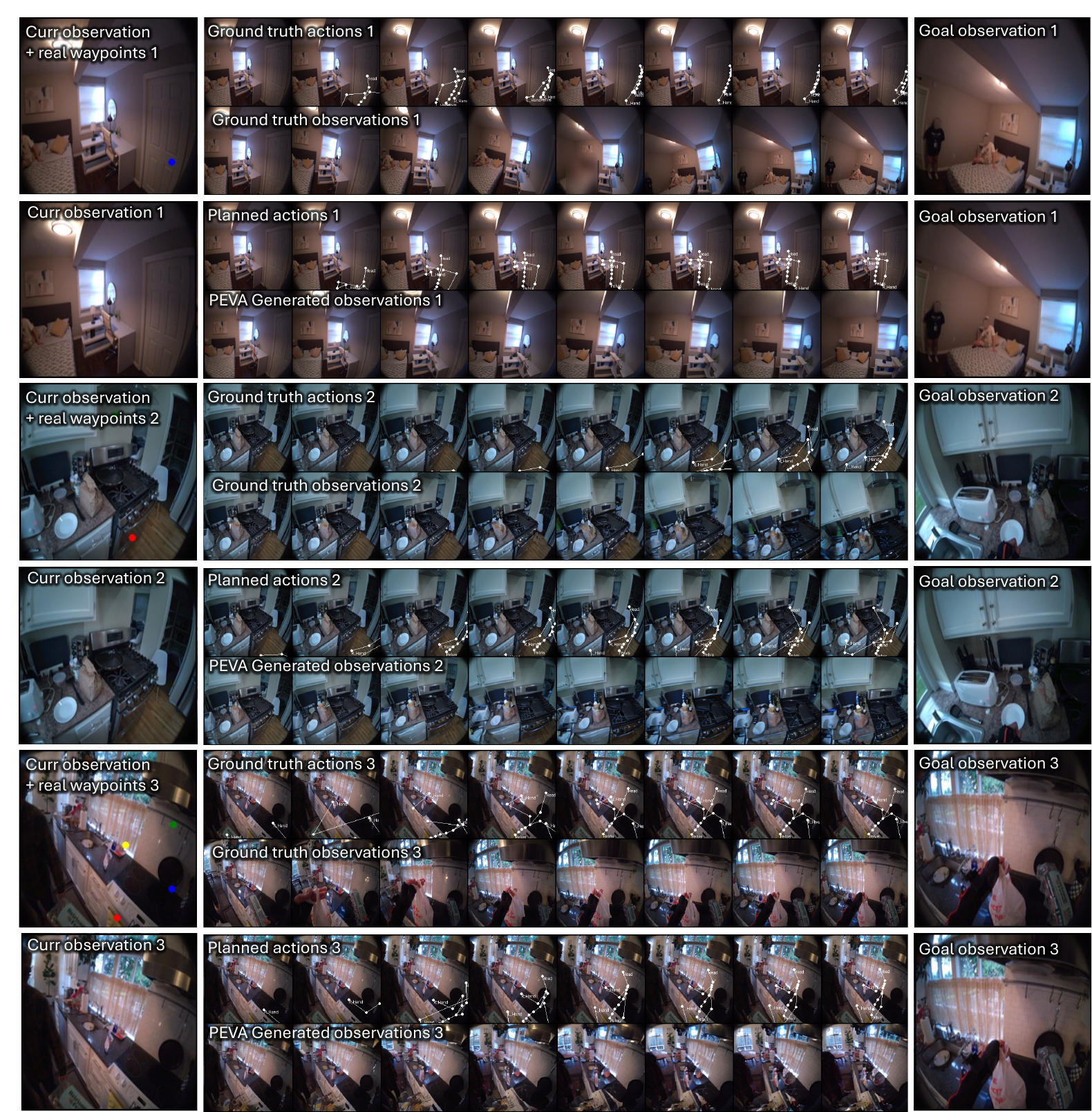}
  \caption{%
    Additional PEVA CEM Planning visualizations without meshes
    }
  \label{fig:peva_planning_vis_skeleton}
\end{figure}
\newpage
\begin{figure}[h]
  \centering
  \includegraphics[width=0.98\linewidth]{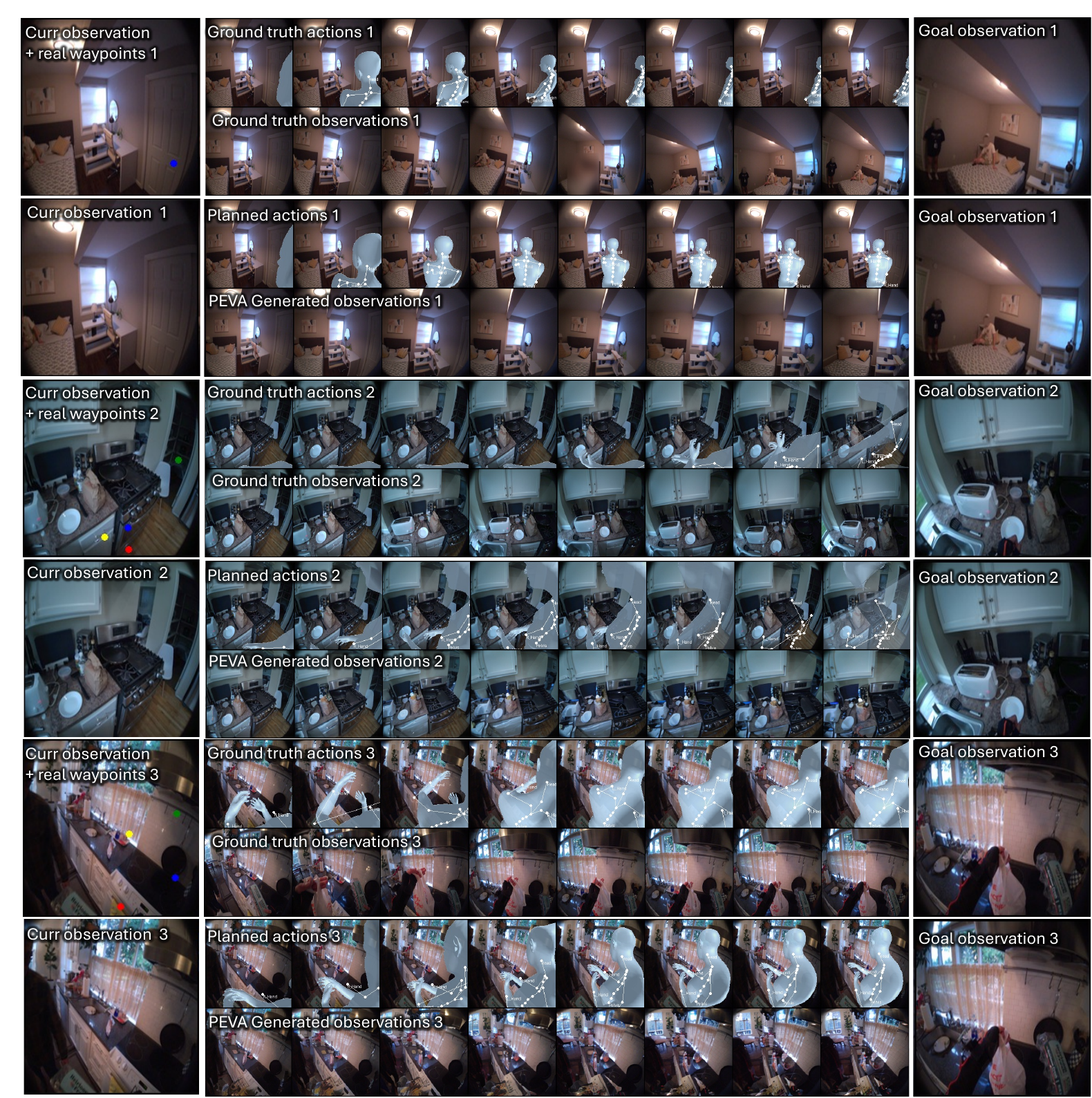}
  \caption{%
    Additional PEVA CEM Planning visualizations
    }
  \label{fig:peva_planning_vis_skinned}
\end{figure}
Note that when planning with PEVA there can be unrealistic body movements. 
See Task 2, where the left shoulder contorted at timesteps 7 and 8.
Also see Task 3 where the torso and right shoulder are in unnatural positions at timesteps 3 through 8.

\newpage
\section{Additional Policy Visualizations}
\label{sec:additional_policy_vis}
\begin{figure}[h]
    \centering
    \includegraphics[width=0.8\linewidth]{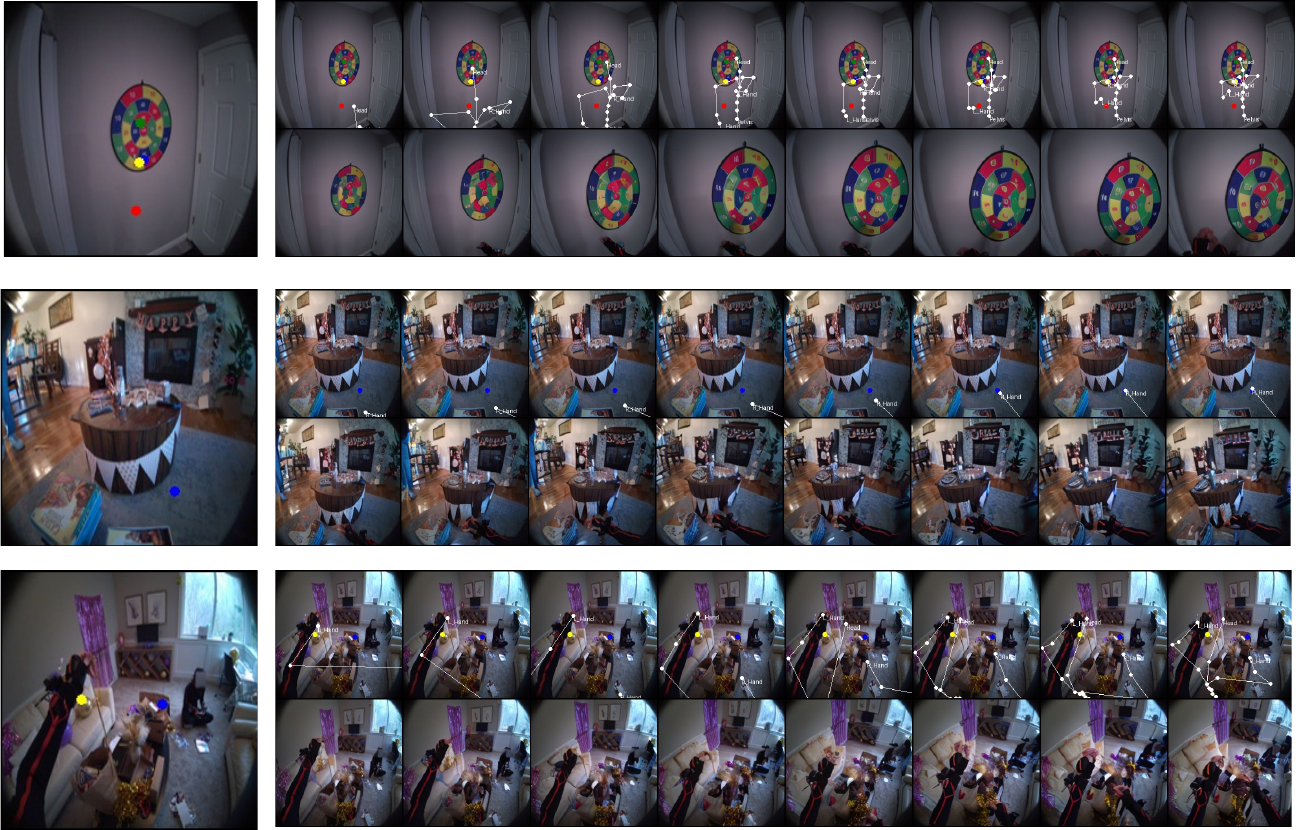}
    \caption{Additional visualizations of waypoint actions and rollout.}
    \label{fig:additional_vis1}
\end{figure}

\begin{figure}[h]
    \centering
    \includegraphics[width=0.8\linewidth]{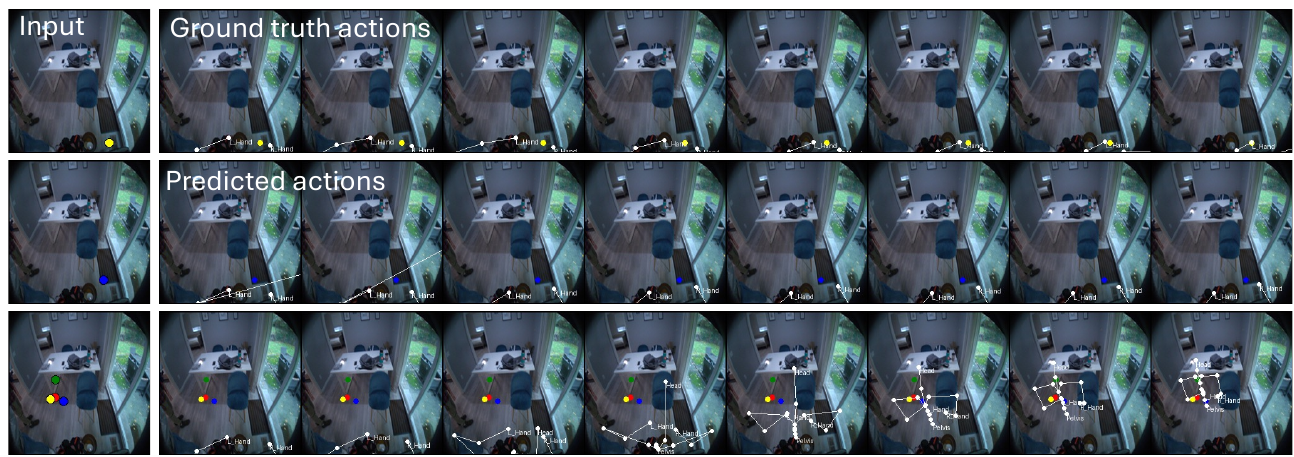}
    \caption{Extra Counterfactual Visualizations 1}
    \label{fig:additional_vis2}
\end{figure}

\begin{figure}[h]
    \centering
    \includegraphics[width=0.8\linewidth]{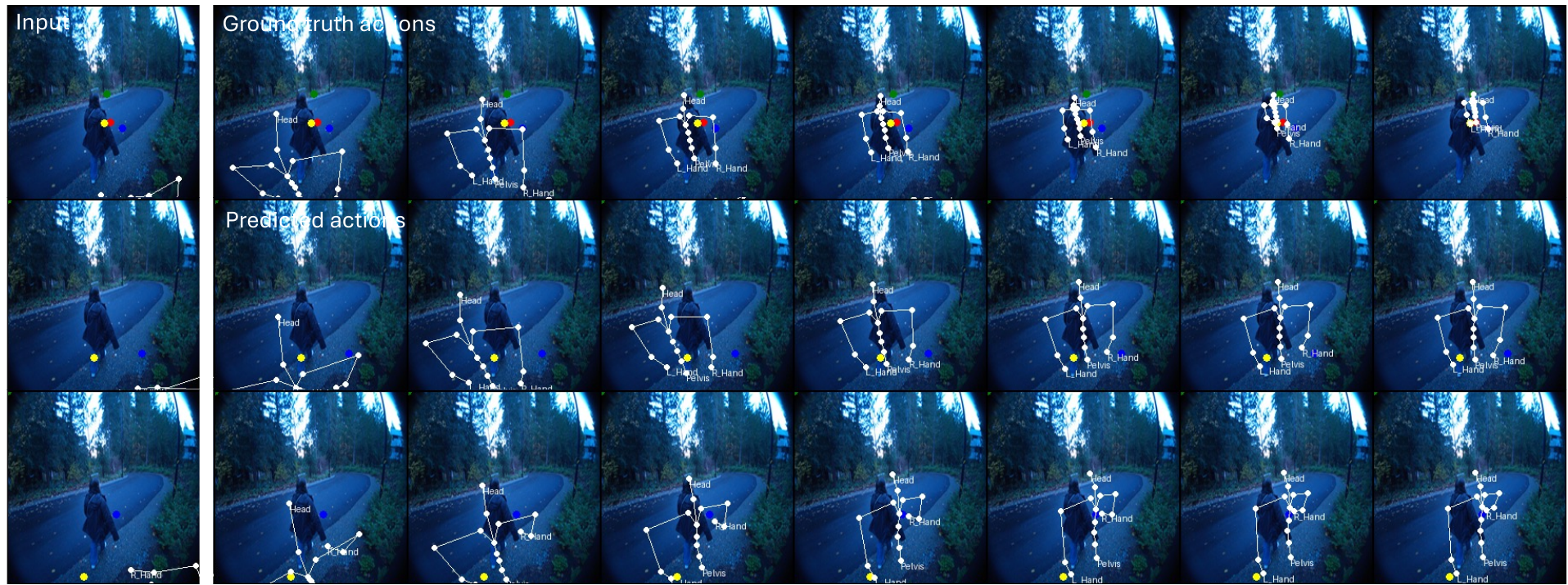}
    \caption{Extra Counterfactual Visualizations 2}
    \label{fig:additional_vis3}
\end{figure}

\end{document}